\documentclass[11pt]{article}
\usepackage{kotex}

\usepackage[final]{acl}

\usepackage{times}
\usepackage{latexsym}
\usepackage{booktabs}
\usepackage{float}
\usepackage{tabularx}
\usepackage{array}
\usepackage{makecell}
\usepackage{amsmath}
\usepackage{amssymb}
\usepackage{multirow}
\usepackage{xcolor}
\usepackage[most]{tcolorbox}
\usepackage{listings}
\usepackage{fancyvrb}

\newcolumntype{Y}{>{\raggedright\arraybackslash}X}

\usepackage[T1]{fontenc}
\usepackage[utf8]{inputenc}
\usepackage{microtype}
\usepackage{inconsolata}
\usepackage{graphicx}
\usepackage{subcaption}

\title{Skill Following: Evaluating Actual Skill Use in Retrieval-Enabled LLM Agents}

\author{Seonghyeon Cho, Chanjun Park\textsuperscript{\textdagger}
\\
\\Soongsil University
\\
   \texttt{chosh040@soongsil.ac.kr, chanjun.park@ssu.ac.kr}
}

\begin{document}
\maketitle

\begingroup
\renewcommand{\thefootnote}{\textdagger}
\footnotetext{Corresponding authors.}
\endgroup

\begin{abstract}
Large Language Model (LLM) agents increasingly rely on external skills, yet standard evaluations obscure whether retrieving these skills actually helps.
Aggregate metrics often compare retrieved versus non-retrieved tasks, introducing severe selection bias and failing to isolate the true effect of skill use.
To measure this actual-use capability—which we formalize as \textit{Skill Following} (SF)—we introduce the \textit{Retrieval-Invoked Actual-Use Effect} (RAE).
RAE computes the same-task outcome difference between matched skill-enabled and skill-disabled executions, conditioned exclusively on tasks where the agent actively retrieved a skill.
Evaluating 17 LLMs across coding and mathematical domains, we uncover a stark evaluation paradox: models frequently show positive aggregate retrieval lift but negative RAE.
On MBPP+, multiple models that appear to benefit system-wide actually harm their own performance on the exact tasks where retrieval occurred.
These findings demonstrate that aggregate averages can create a misleading illusion of tool-use proficiency, whereas RAE directly measures whether the retrieval-to-answer pipeline genuinely rescues more outcomes than it harms.
\end{abstract}

\begin{figure}[t]
    \centering
    \includegraphics[width=\linewidth]{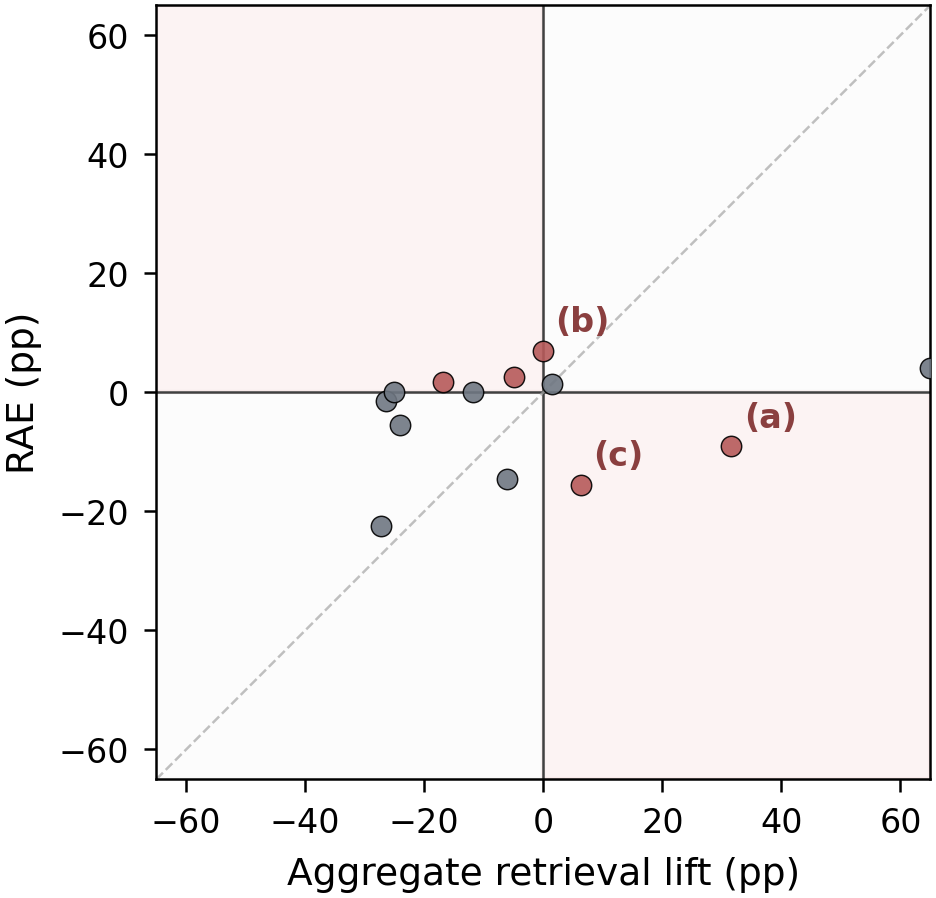}
    \caption{
        Aggregate retrieval lift and RAE on MBPP+. The x-axis compares skill-enabled accuracy between retrieval-returned and retrieval-skipped task sets, whereas the y-axis compares paired skill-enabled and skill-disabled outcomes on the same retrieval-returned tasks. Shaded off-diagonal quadrants indicate sign disagreement; labels (a)–(c) mark the cases discussed in the text.
    }
    \label{fig:fig1}
\end{figure}

\section{Introduction}\label{sec:1}
Large language model (LLM) agents are increasingly deployed within external procedural environments, utilizing tools, memory, and skill libraries to navigate and solve complex tasks~\cite{schick2023toolformerlanguagemodelsteach,wu2024avataroptimizingllmagents,shen2024smallllmsweaktool,shi2024learning}.
As these models transition from standalone text predictors to retrieval-enabled, interactive agents, the evaluation paradigms used to assess their efficacy must evolve correspondingly.
However, measuring the true utility of external skills remains fundamentally challenging. Existing evaluation protocols typically answer one of two broad, system-level questions: whether providing an agent with access to a skill environment improves average performance across an entire benchmark, or whether tasks where skills were retrieved yield higher accuracy than tasks where they were skipped. While these aggregate metrics provide a high-level summary of system behavior, they leave a critical, narrower question systematically unanswered: \textit{when an agent actually chooses to retrieve and inject a skill into its context, does that specific external knowledge genuinely improve the outcome of that precise task?}

We argue that aggregate benchmark averages inherently obscure the true causal utility of skill retrieval, often creating a misleading illusion of capability. Whole-benchmark metrics dilute the impact of external tools by heavily weighting tasks where no skills were ever fetched. More problematically, aggregate retrieved-vs-skipped comparisons suffer from severe task-selection bias; they compare two self-selected subsets of tasks that may fundamentally differ in inherent difficulty, prompt complexity, or skill-library relevance. Because autonomous retrieval is merely the preamble of tool use—requiring the LLM to subsequently receive, interpret, logically ground, and synthesize the returned content into a valid final answer—evaluating this complete pipeline demands a more rigorous methodology. We formalize this multi-step, retrieval-to-answer cognitive process as \textit{Skill Following} (SF). To accurately evaluate SF without the confounding variables of task self-selection, the measurement must be explicitly conditioned on the actual occurrence of retrieval, utilizing a strictly paired experimental design.

To clearly illustrate the severity of this evaluation gap, Figure~\ref{fig:fig1} contrasts the conventional aggregate retrieval lift against our proposed same-task metric on the MBPP+ benchmark. The x-axis represents the aggregate lift—the standard approach that suggests many modern LLMs, such as Gemini-2.5-Flash-lite and DeepSeek-V3.2, achieve substantial performance gains when skills are retrieved.

However, when we isolate the exact subset of tasks where these models successfully retrieved a skill and compare their outcomes strictly against their own skill-disabled baselines on those identical tasks (the y-axis), a starkly different reality emerges. The shaded quadrants denote regions of critical \textit{sign disagreement}: models positioned here exhibit a positive aggregate lift that statistically masks a negative actual-use effect. This paradox reveals that what appears to be a systemic improvement is often just a byproduct of the model selectively retrieving skills on inherently easier tasks, while the actual integration of those skills frequently disrupts the model's reasoning, ultimately rescuing fewer outcomes than it harms.

To correct this methodological blind spot and isolate the actual-use capability of LLMs, we introduce the \textit{Retrieval-Invoked Actual-Use Effect} (RAE). RAE is a protocol-conditional, same-task outcome metric computed exclusively on the tasks where retrieval was actively invoked and a skill was successfully returned in the skill-enabled execution. By mandating a paired skill-enabled versus skill-disabled comparison, RAE strips away population-level confounding factors and directly measures whether the retrieval-invoked SF chain operates as a genuine cognitive aid or a distracting bottleneck.

In summary, this paper makes the following primary contributions:
\begin{itemize}
    \item \textbf{Identifying the Actual-Use Gap in Agent Evaluation:} We conceptualize skill integration as the full \textit{Skill Following} (SF) chain and mathematically demonstrate that standard aggregate metrics fail to isolate the true task-level impact of retrieved skills due to selection bias.

    \item \textbf{Introducing the RAE Metric:} We propose the Retrieval-Invoked Actual-Use Effect (RAE), establishing a rigorous, paired-evaluation protocol that conditions outcome measurement specifically on active retrieval events.

    \item \textbf{Exposing Pervasive Metric Reversals across LLMs:} Across an extensive evaluation of 17 diverse LLMs spanning coding and mathematical reasoning domains, we reveal that aggregate conclusions frequently reverse under RAE. Crucially, our comprehensive diagnostic analyses prove that mere context exposure to retrieved skills is vastly insufficient for successful Skill Following, challenging current assumptions about LLM tool-use capabilities.
\end{itemize}

\section{Related Work}\label{sec:2}
LLM agents are increasingly evaluated not merely as standalone text predictors, but as complex systems integrating external tools, memory, and retrieval mechanisms. Benchmarks such as AgentBench~\cite{liu2024agentbench}, AgentBoard~\cite{ma2024agentboard}, WebArena~\cite{zhou2024webarena}, OSWorld~\cite{xie2024osworld}, TheAgentCompany~\cite{xu2026theagentcompany}, and $\tau$-bench~\cite{yao2024tau} measure end-to-end agent behavior in interactive or environment-grounded settings. While these benchmarks provide crucial system-level evaluations, their primary metrics focus on overall task success or step-wise process diagnostics. Consequently, they do not isolate whether the injection of retrieved external content directly changed the final outcome of that exact task via paired comparisons.

Benchmarks specifically targeting external skill access provide a closer point of comparison. Datasets such as SkillsBench~\cite{li2026skillsbench}, SWE-Skills-Bench~\cite{han2026swe}, and SkillLearnBench~\cite{zhong2026skilllearnbench} employ paired designs to contrast skill-enabled conditions against no-skill or human-skill baselines. Although this paired approach effectively controls for task identity, their reported effects are still predominantly aggregated over the entire evaluation set. In contrast, RAE precisely isolates the actual-use component of the Skill Following chain by conditioning the paired comparison strictly on the subset of tasks where retrieval was actively invoked and content was returned.

A complementary line of research investigates autonomous retrieval, memory, and skill reuse within agents, including systems like Voyager~\cite{wang2023voyager}, Agent KB~\cite{tang2025agent}, ExpeL~\cite{zhao2024expel}, Reflexion~\cite{shinn2023reflexion}, and CRITIC~\cite{gou2024critic}. While these frameworks preserve and demonstrate agent-initiated use of external artifacts, they typically do not systematically pair each retrieval-invoked execution with a corresponding skill-disabled baseline to measure the net performance shift. Furthermore, process-level benchmarks such as BFCL~\cite{patil2025berkeley}, RAGAS~\cite{es2024ragas}, and IFEval~\cite{zhou2023instruction} diagnose specific sub-stages like tool calling, retrieval quality, or grounding. RAE is fully complementary to these diagnostics: rather than replacing stage-level metrics, it evaluates the end-to-end impact of the retrieval-invoked chain, strictly asking whether the complete process ultimately rescued more same-task outcomes than it harmed.

\section{Skill Following}\label{sec:method}

In this section, we establish the formal framework of paired outcome metrics required to rigorously evaluate the task-level efficacy of Skill Following (SF). Conceptualizing SF as the complete retrieval-to-answer cognitive chain introduced in Section~\ref{sec:1}, we define these metrics through strictly paired comparisons between skill-enabled and skill-disabled executions on identical tasks.

\begin{table*}[t]
  \centering
  \small
  \setlength{\tabcolsep}{6pt}
  \begin{tabular}{p{0.22\linewidth} p{0.25\linewidth} c p{0.34\linewidth}}
    \toprule
    \textbf{Metric} & \textbf{Conditioning set} & \textbf{Paired?} & \textbf{Comparison} \\
    \midrule
    Aggregate retrieval lift
    & SE retrieved vs.\ SE non-retrieved
    & No
    & Mean accuracy difference between two task subsets within the skill-enabled execution \\
    \midrule
    OAE
    & $S_{\mathrm{all}}$
    & Yes
    & SE--SD outcome difference over all evaluated tasks \\
    \midrule
    RAE (ours)
    & $S_{\mathrm{call}}$
    & Yes
    & SE--SD outcome difference over retrieval-invoked tasks \\
    \bottomrule
  \end{tabular}
  \caption{Structural taxonomy of aggregate retrieval lift, OAE, and RAE. OAE and RAE are paired, same-task quantities, whereas aggregate retrieval lift compares self-selected task subsets within the skill-enabled execution.}
  \label{tab:metrics_comparison}
\end{table*}

\subsection{Paired Execution Design}\label{sec:3.1}

To construct a controlled comparison, each task $t$ within the evaluation corpus undergoes two distinct execution paths. The \textit{skill-enabled} (SE) execution grants the agent full access to the skill library and retrieval interface, whereas the \textit{skill-disabled} (SD) baseline restricts this access. Let $y^{+}_t,y^{-}_t \in \{0,1\}$ denote the binary correctness outcomes of the SE and SD executions, respectively. The paired outcomes $(y^{+}_t,y^{-}_t) \in \{0,1\}^2$ map each task into one of four distinct transition categories (Table~\ref{tab:on_off}). This strictly paired design anchors the evaluation to fixed task identities, inherently neutralizing the confounding effects of cross-population task difficulty.

\begin{table}[H]
    \centering
    \small
    \begin{tabular}{lcc}
        \toprule
        & \textbf{SE correct} & \textbf{SE wrong} \\
        \midrule
        \textbf{SD correct}
        & concordant pass
        & harmful flip $(c)$ \\
        \textbf{SD wrong}
        & helpful flip $(b)$
        & concordant fail\\
        \bottomrule
    \end{tabular}
    \caption{Outcome matrix for paired skill-enabled (SE) and skill-disabled (SD) executions. $b$ denotes the number of tasks that failed in SD but succeeded in SE, whereas $c$ denotes the converse.}
    \label{tab:on_off}
\end{table}

\subsection{Overall and Retrieval-Invoked Effects}\label{sec:metric_definitions}

Because autonomous retrieval is a selective process, the SE execution will successfully fetch skills for certain tasks but not others. Consequently, we delineate two distinct paired metrics: the \textit{Overall Skill-Access Effect} (OAE), computed over the entire evaluation corpus, and the \textit{Retrieval-Invoked Actual-Use Effect} (RAE), strictly conditioned on tasks where retrieval was actively invoked and a skill was returned.

\paragraph{Overall skill-access effect.}
Let $S_{\mathrm{all}}$ denote the full evaluated task set. Equation~\ref{eq:oae} defines OAE as the macroscopic, paired outcome difference across the entire benchmark:
\begin{equation}
    \begin{aligned}
        \mathrm{OAE} = \Delta_{\mathrm{all}}
        &= \frac{1}{|S_{\mathrm{all}}|}\sum_{t \in S_{\mathrm{all}}}(y_t^{+} - y_t^{-}) \\
        &= \frac{\mathrm{all}_b - \mathrm{all}_c}{n_{\mathrm{pairs}}},
    \end{aligned}
    \label{eq:oae}
\end{equation}
where $n_{\mathrm{pairs}}=|S_{\mathrm{all}}|$, and $\mathrm{all}_b$ and $\mathrm{all}_c$ represent the total helpful and harmful flips over the full dataset.

OAE quantifies the systemic impact of exposing the agent to a skill environment. It computes the average outcome shift across all paired executions, deliberately encompassing tasks where the SE execution bypassed retrieval entirely. While OAE accurately captures the global utility of skill access, it fundamentally dilutes the signal of actual skill usage.

\paragraph{Retrieval-invoked actual-use effect}
To distill the genuine impact of active skill utilization, we strictly condition our measurement on the subset of tasks where the SE execution invoked retrieval and successfully acquired at least one skill. Let $S_{\mathrm{call}}$ be this retrieval-invoked subset, with size $n_{\mathrm{cond}} = |S_{\mathrm{call}}|$. Equation~\ref{eq:rae} defines RAE as the paired outcome difference isolated to this critical subset:
\begin{equation}
    \begin{aligned}
        \mathrm{RAE} = \Delta_{\mathrm{cond}}
        &= \frac{1}{|S_{\mathrm{call}}|}\sum_{t \in S_{\mathrm{call}}}(y_t^{+} - y_t^{-}) \\
        &= \frac{\mathrm{cond}_b - \mathrm{cond}_c}{n_{\mathrm{cond}}}.
    \end{aligned}
    \label{eq:rae}
\end{equation}

Crucially, unlike aggregate retrieval lift---which merely contrasts self-selected retrieved and non-retrieved populations within the SE condition---RAE is an unconfounded, same-task metric. It directly addresses whether the retrieval-invoked segment of the SF chain successfully improved outcomes relative to the mathematically matched SD baselines. Table~\ref{tab:metrics_comparison} delineates the structural distinctions between aggregate retrieval lift, OAE, and RAE.

\paragraph{Relationship between OAE and RAE}
RAE operates as a conditional sub-component of the broader skill-access effect.
Let $S_{\mathrm{skip}}=S_{\mathrm{all}}\setminus S_{\mathrm{call}}$ denote the subset of tasks where retrieval was circumvented, with size $n_{\mathrm{skip}}=|S_{\mathrm{skip}}|$.
We define the paired effect over this uninvoked complement as:
\begin{equation}
    \Delta_{\mathrm{skip}}
    =
    \frac{\mathrm{skip}_b-\mathrm{skip}_c}{n_{\mathrm{skip}}},
    \label{eq:skip_effect}
\end{equation}
where $\mathrm{skip}_b=\mathrm{all}_b-\mathrm{cond}_b$ and $\mathrm{skip}_c=\mathrm{all}_c-\mathrm{cond}_c$. Combining the retrieval-invoked and skipped components yields a full decomposition of the benchmark effect:
\begin{equation}
    \mathrm{OAE}
    =
    \frac{n_{\mathrm{cond}}}{n_{\mathrm{pairs}}}\mathrm{RAE}
    +
    \frac{n_{\mathrm{skip}}}{n_{\mathrm{pairs}}}\Delta_{\mathrm{skip}}.
    \label{eq:oae_decomp}
\end{equation}

Equation~\ref{eq:oae_decomp} demonstrates that RAE does not supersede OAE; rather, it isolates the critical mechanism of actual usage.
OAE answers the overarching system-design question, whereas RAE targets the precise actual-use inquiry.
RAE is reported for each evaluated model--skill-pool--retrieval-policy configuration, whose retrieval-returned subset is system-specific.
We therefore report RAE alongside its granular helpful and harmful transition counts; comprehensive details regarding statistical tests, confidence intervals, and edge-case handling are provided in Appendix~\ref{app:metric_details}.

\section{Experiment Setup}\label{sec:4}

Our experimental design is structured to address four primary objectives. First, we investigate whether aggregate retrieval lift and RAE yield conflicting conclusions under an identical skill-enabled protocol. Second, we analyze whether aggregate retrieval lift is confounded by retrieval-induced task selection. Third, we validate the consistency of our findings across multiple task partitions, a secondary coding benchmark, and an entirely different domain (mathematical reasoning). Finally, we conduct transition-level diagnostics and skill-content controls to rigorously bound the interpretation of negative RAE outcomes.

\subsection{Models and Datasets}\label{sec:4.1}
We evaluate our proposed measurement framework across a comprehensive panel of 17 diverse LLMs. This panel spans various parameter scales and includes both closed-source APIs and open-weight models to ensure broad generalizability (see Appendix~\ref{app:models} for complete model details).

For our primary analysis, we sample 80 tasks from the MBPP+ benchmark~\cite{evalplus} using seed 42. To verify cross-benchmark consistency, we evaluate the models on HumanEval+~\cite{chen2021evaluating}. Furthermore, to confirm that the observed phenomena are not strictly domain-dependent, we perform a cross-domain replication using 80-task partitions of Math500~\cite{hendrycksmath2021,lightman2024let}.

\subsection{Skill Environment and Retrieval Mechanism}\label{sec:4.2}
All analyses operate on fixed, procedural skill libraries tailored to the respective domains. The \textit{Coding skill pool} comprises 9 procedural coding skills used for both MBPP+ and HumanEval+. The \textit{Math skill pool} comprises 8 procedural skills used for the Math500 evaluation. Appendix~\ref{app:skill_pools} summarizes both skill pools, and Appendix~\ref{app:skill_and_prompts} reports the prompts and tool schema.

Each skill is formatted as a structured Markdown file (\texttt{SKILL.md}), containing lightweight frontmatter (name, description, signature) and a detailed body (application conditions, code snippets, invariants, and self-tests). The retrieval system indexes the skill descriptions and Markdown bodies using BM25~\cite{robertson2009probabilistic}. When an agent autonomously invokes the \texttt{search\_skills(query)} function, the full texts of the top-3 retrieved skills are systematically appended to its working context. Agents are permitted up to three search invocations per task.

\subsection{Paired Execution Protocol}\label{sec:4.3}
To compute the same-task metrics detailed in \S\ref{sec:method}, each task $t$ undergoes two strictly controlled, parallel executions. In the SE condition, the LLM receives a system prompt that explicitly defines and grants access to the \texttt{search\_skills(query)} tool. In the SD baseline condition, the model receives an identical prompt, but with the tool definition entirely removed. Both execution paths share the exact same task prompt, generation seed, and decoding configuration, ensuring that any outcome variance is strictly attributable to the skill access and subsequent retrieval chain.

\subsection{Diagnostic Controls and Annotation}\label{sec:4.4}
To interpret the mechanistic drivers of negative RAE, we design two supplementary diagnostic setups for the strongest coding reversal models (DeepSeek-V3.2 and Gemini-2.5-Flash-lite).

\paragraph{Skill-Content Controls.} We run identical evaluations on MBPP+ under five perturbed environments: \textit{Normal} (the standard skill library), \textit{Schema-Empty} (tool is available but returns no usable skills), \textit{Filler-Dummy} (matched non-informative text), \textit{Random-skills} (retrieval returns structurally valid but completely irrelevant skills), and \textit{Corrupted} (retrieved content is intentionally misleading or mathematically inconsistent).

\paragraph{Adherence Annotation}
For transition-level diagnosis, a GPT-5.5 annotator annotates the retrieval-invoked executions.
The annotator evaluates the task prompt, retrieval query, retrieved skill excerpts, and the final SE answer to assign a primary behavioral label: \textit{appropriate adherence, ignored/independent, misapplied/overapplied, format/interface failure,} or \textit{unclear}.
The annotator is blinded to both the SD answers and the underlying correctness labels; the resulting post-hoc labels provide descriptive diagnostics of the observed SE traces, remain separate from the RAE computation, and are evaluated through a stratified human audit reported in Appendix~\ref{app:human_audit}.

\section{Results}\label{sec:results}
In this section, we investigate whether aggregate retrieval lift and our proposed RAE yield consistent evaluations of LLM tool-use capabilities. Through a comprehensive analysis across coding and mathematical benchmarks, we demonstrate that standard aggregate metrics frequently present a misleading illusion of performance. We structurally unpack this discrepancy by diagnosing task-selection bias, cross-domain consistency, and the precise mechanistic failures that occur once external content enters the model's context.

\subsection{Do Aggregate Metrics and RAE Agree?}\label{sec:agg_can_reverse}
We first establish the prevalence of metric disagreement. Table~\ref{tab:main_results} reports the coding-domain results, contrasting aggregate retrieval lift with RAE.

\paragraph{Sign Disagreements Across Model and Benchmark}
Sign reversals—where models exhibit a positive aggregate lift but a negative actual-use effect (RAE)—are alarmingly common. Across MBPP+, Table~\ref{tab:main_results} shows repeated sign disagreements among reportable model configurations.

On HumanEval+, 3 out of 13 models exhibit the same paradox. For instance, Gemini-2.5-Flash-lite consistently shows positive aggregate lift across all three MBPP+ partitions, yet its RAE remains strictly negative. These findings highlight a critical vulnerability in current evaluation standards: seemingly beneficial skill access at the benchmark level frequently masks actual performance degradation on the specific tasks where skills are deployed.

\begin{table*}[t]
    \centering
    \small
    \setlength{\tabcolsep}{6pt}
    \begin{tabular}{l cccc}
        \toprule
        \textbf{Model}
        & \textbf{MBPP+ s42}
        & \textbf{MBPP+ s43}
        & \textbf{MBPP+ s44}
        & \textbf{HumanEval+ s42} \\
        \midrule
        Qwen3-8B & $-26.4$ / $-1.5$ & $-0.5$ / $-9.2$ & $-26.6$ / $-1.6$ & \textbf{$-10.5$ / $+5.7$} \\
        Qwen3.5-9B & $-11.8$ / $+0.0$ & \textbf{$+26.1$ / $-3.2$} & $-36.4$ / $-4.1$ & \textbf{$-5.4$ / $+7.2$} \\
        Gemini-2.5-Flash-lite & \textbf{$+6.2$ / $-15.6$} & \textbf{$+16.2$ / $-15.9$} & \textbf{$+4.4$ / $-22.2$} & $-0.1$ / $-21.1$ \\
        Mistral-S-24B & $-27.2$ / $-22.5$ & \textbf{$+10.8$ / $-9.0$} & $-37.9$ / $-45.5$ & $-13.9$ / $-38.1$ \\
        \midrule
        GPT-4o-mini & $-24.0$ / $-5.5$ & \textbf{$+11.4$ / $-3.4$} & $-23.2$ / $-8.2$ & $-2.5$ / $-1.4$ \\
        Claude-3.5-H & \textbf{$-5.0$ / $+2.5$} & $-1.0$ / $-2.1$ & $-6.6$ / $-3.8$ & $-2.7$ / $-7.7$ \\
        Claude-4.5-H & $+1.4$ / $+1.4$ & $+21.5$ / $+0.0$ & $+48.0$ / $+2.7$ & $+4.5$ / $+0.0$ \\
        Qwen3-32B & \textbf{$-16.9$ / $+1.8$} & $-27.9$ / $-6.0$ & $-12.4$ / $-8.5$ & $-11.2$ / $-30.1$ \\
        \midrule
        GLM4-32B & $-25.1$ / $+0.0$ & \textbf{$+1.8$ / $-1.4$} & $-13.9$ / $-6.8$ & $-2.8$ / $-5.6$ \\
        Command-R & $-6.2$ / $-14.5$ & $-36.1$ / $-24.1$ & $-33.3$ / $-26.7$ & $-5.8$ / $+0.0$ \\
        Llama-3.3-70B & $+69.8$ / $+4.2$ & \textbf{$+44.2$ / $-2.0$} & $+60.0$ / $+7.5$ & $+61.5$ / $+7.1$ \\
        Qwen3-235B & \textbf{$-0.1$ / $+7.0$} & $-21.1$ / $+0.0$ & \textbf{$-27.1$ / $+3.7$} & \textbf{$-7.0$ / $+15.5$} \\
        DeepSeek-V3.2 & \textbf{$+31.6$ / $-9.1$} & $-11.5$ / $-15.4$ & \textbf{$+46.2$ / $-23.1$} & $-17.1$ / $-23.0$ \\
        \bottomrule
    \end{tabular}

    \caption{
        Main coding-domain aggregate-vs-RAE comparison. Each cell reports aggregate retrieval lift / RAE in percentage points. Bold entries indicate opposite signs between the unpaired retrieved-vs-skipped aggregate and the retrieval-invoked same-task metric. The table includes cells satisfying the reporting filter; the full 17-model panel, including non-reportable cells, is provided in Appendix~\ref{app:full_coding_results}.
    }
    \label{tab:main_results}
\end{table*}

\begin{figure}[H]
  \centering
  \includegraphics[width=\linewidth]{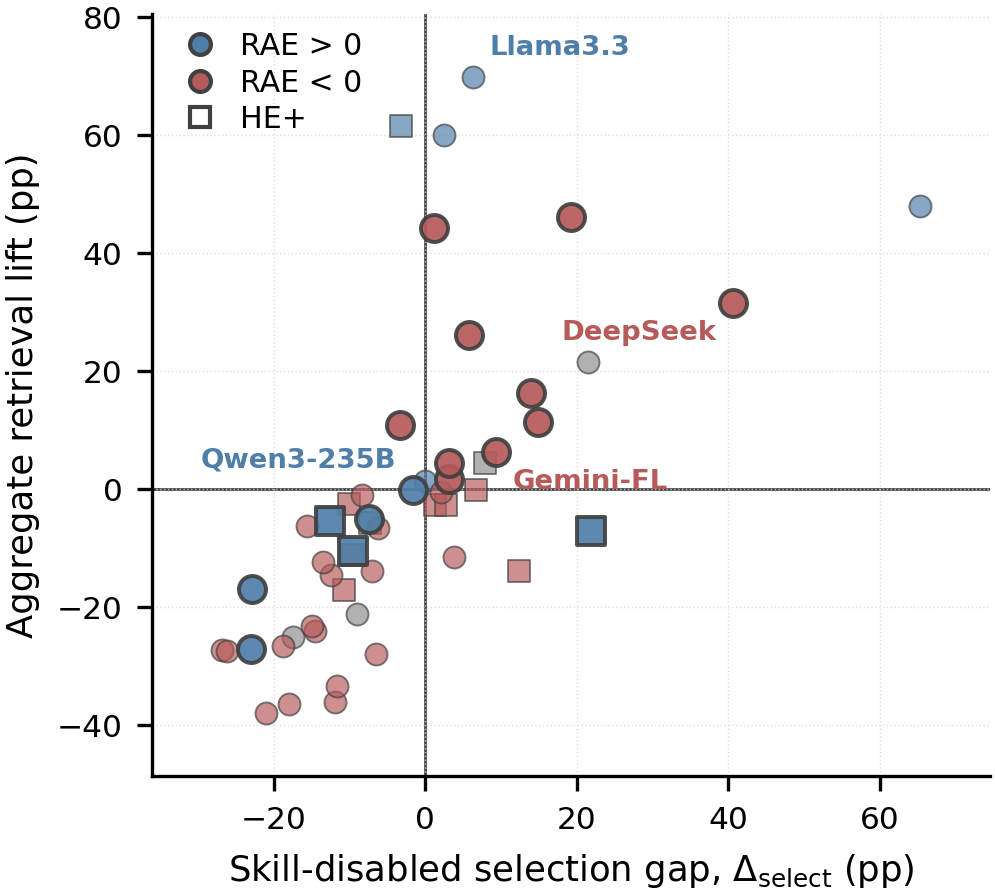}
    \caption{Autonomous retrieval induces severe task-selection bias. The x-axis reports the skill-disabled selection gap ($\Delta_{\mathrm{select}}$) between tasks that did and did not return a skill. The y-axis reports aggregate retrieval lift within the skill-enabled execution. Each point represents a reportable model cell, colored by the sign of its RAE. The prevalent nonzero selection gaps mathematically demonstrate that aggregate retrieval lift compares inherently mismatched task populations, failing to isolate the true causal effect of skill usage.}
  \label{fig:selection_gap}
\end{figure}

\subsection{What Drives the Divergence Between Aggregate Lift and Actual Use?}\label{sec:agg_ref_ret_sel}
To understand \textit{why} aggregate retrieval lift diverges so sharply from RAE, we must analyze the task populations being compared. Aggregate lift evaluates retrieved versus non-retrieved tasks within the skill-enabled execution. However, these subsets are not randomized; they are self-selected by the LLM's own retrieval policy.

To quantify this underlying discrepancy, we introduce the skill-disabled selection gap ($\Delta_{\mathrm{select}}$). Let $S_{\mathrm{ret}}$ and $S_{\mathrm{noret}}$ denote the subsets of tasks that did and did not return a skill, respectively. We measure their inherent difficulty using the paired SD baseline:
\begin{equation}
    \Delta_{\mathrm{select}}
    =
    \mathrm{Acc}^{-}(S_{\mathrm{ret}})
    -
    \mathrm{Acc}^{-}(S_{\mathrm{noret}})
\end{equation}

\paragraph{Selection Bias Confounds Aggregate Lift}
As illustrated in Figure~\ref{fig:selection_gap}, autonomous retrieval induces substantial task-selection differences ($\Delta_{\mathrm{select}} \neq 0$). This provides the mechanistic explanation for the metric disagreement: aggregate retrieval lift effectively compares two distinct, self-selected task populations that already differ in baseline difficulty or prompt complexity. RAE, by contrast, neutralizes this bias by strictly comparing the same retrieval-invoked tasks against their paired SD executions.

\subsection{Does the Evaluation Discrepancy Persist Beyond Benchmarks?}\label{sec:disagre_across}
To confirm that this paradox is not an artifact of a single benchmark, we extend our analysis to alternative datasets and domains.

\paragraph{Cross-Benchmark Consistency}
The aggregate-vs-RAE disagreement is not confined to MBPP+. On HumanEval+, 3 out of 13 reportable models show sign disagreement between aggregate retrieval lift and RAE. Unlike the dominant MBPP+ reversal emphasized in the introduction, these HumanEval+ disagreements occur in the opposite direction: aggregate retrieval lift is negative while RAE is positive for Qwen3-8B, Qwen3.5-9B, and Qwen3-235B. This indicates that aggregate retrieved-vs-skipped comparisons can misestimate retrieval-invoked actual use in either direction, either overstating or understating the effect of SF. Appendix~\ref{app:cross_benchmark_corr} reports descriptive MBPP+--HumanEval+ correlation analyses, which we use only as diagnostic context rather than as benchmark-invariant model rankings.

\begin{figure}[t]
  \centering
  \includegraphics[width=\linewidth]{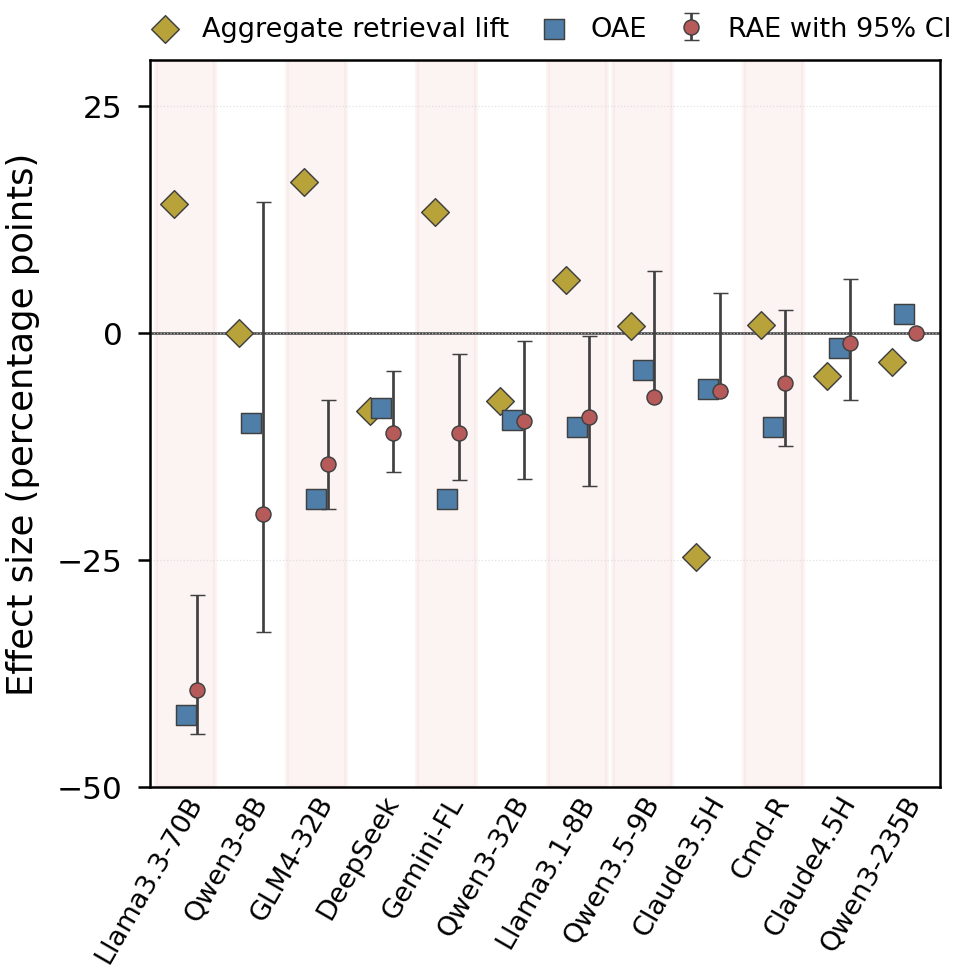}
  \caption{Math500 cross-domain results. Comparing aggregate retrieval lift, OAE, and RAE reveals that the metric disagreement generalizes beyond coding. Models with highly positive aggregate lift collapse into severely negative actual-use effects (RAE).}
  \label{fig:fig2}
\end{figure}

\paragraph{Cross-Domain Results on Math500}
Figure~\ref{fig:fig2} shows that the disagreement between aggregate retrieval lift and RAE also appears in the mathematical reasoning domain. On Math500, Llama-3.3-70B obtains a $+14.2$ pp aggregate lift but a $-39.4$ pp RAE under the paired retrieval-invoked comparison. Gemini-2.5-Flash-lite shows a similar pattern ($+13.2$ pp aggregate lift vs. $-11.0$ pp RAE). These results indicate that the aggregate-vs-RAE disagreement is not limited to coding benchmarks. Full Math500 results are reported in Appendix~\ref{app:math500}.

\subsection{Does Higher Retrieval Coverage Translate to Better Downstream Integration?}\label{sec:inv_is_not_sf}
\begin{figure}[t]
  \centering
  \includegraphics[width=\linewidth]{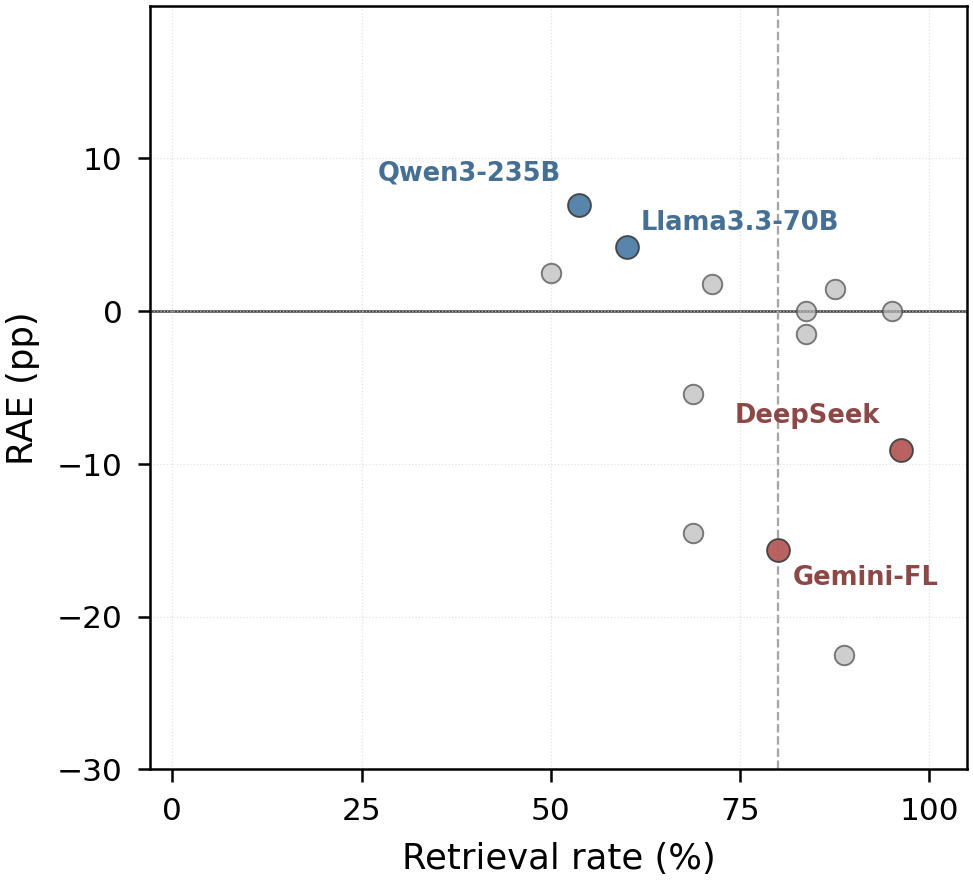}
  \caption{Retrieval coverage versus RAE on MBPP+ seed 42. The horizontal line marks zero RAE. Multiple high-retrieval models exhibit negative RAE, proving that successful retrieval invocation (exposure) does not reliably translate to successful SF (utilization).}
  \label{fig:fig4}
\end{figure}

A prevailing assumption in tool-use evaluation is that higher retrieval coverage naturally translates to improved downstream performance. We examine this assumption by mapping retrieval coverage against RAE (Figure~\ref{fig:fig4}). The results indicate that several models with high retrieval rates nevertheless yield negative actual-use effects. To unpack this discrepancy, the corresponding stage-level diagnostic definitions and values are detailed in Appendix~\ref{app:stage_diagnostics}.

\paragraph{Retrieval is not Following}
If mere exposure to external skills were sufficient, models with higher retrieval coverage would consistently show positive RAE. Instead, Figure~\ref{fig:fig4} indicates that several high-coverage models still exhibit negative RAE. As detailed in the stage-diagnostic analysis (Appendix~\ref{app:stage_diagnostics}), there is a notable gap between fetching a skill and successfully applying it: even when models retrieve skills at a high rate, their structural adherence to the provided content and effective success rates remain substantially lower. This disconnect highlights the distinction between \textit{exposure} and \textit{actual use}. Invoking a search function and receiving returned text does not guarantee that the LLM will correctly interpret, align, and incorporate that content into a valid final answer.

\begin{table*}[t]
  \centering
  \small
  \begin{tabular}{lrr}
      \toprule
      \textbf{Control condition} & \textbf{DeepSeek-V3.2} & \textbf{Gemini-2.5-Flash-lite} \\
      \midrule
      Normal       & $-15.9$ & $-17.9$ \\
      Schema-Empty & --      & -- \\
      Filler-Dummy & $-10.2$ & $-5.0$ \\
      Random-skills & $-15.1$ & $-15.4$ \\
      Corrupted    & $-9.1$  & $-17.2$ \\
      \bottomrule
  \end{tabular}
  \caption{Skill-content controls reporting pooled RAE in percentage points. Schema-Empty returns no usable skill, rendering RAE mathematically undefined ($n_{\mathrm{cond}}=0$). Crucially, RAE remains strictly negative across all conditions where \textit{any} content is returned.}
  \label{tab:content_controls}
\end{table*}

\subsection{Mechanistic Failures: How and Where Does the SF Chain Break Down?}\label{sec:trans_diag_sf}
To diagnose how SF breaks down, we analyze \textit{harmful transitions}—cases where the SD execution succeeds, but the SE execution retrieves a skill and subsequently fails. Using a GPT-5.5 annotator, we annotated 45 harmful transitions for DeepSeek-V3.2 and 40 for Gemini-2.5-Flash-lite across three MBPP+ seeds.

\paragraph{Ignored or Independent Use of Retrieved Skills}
The most frequent failure mode is \textit{ignored/independent}. In these cases, the SE execution retrieves a skill, but the final answer shows no clear evidence that the returned content was incorporated into the solution. DeepSeek-V3.2 also shows \textit{format/interface failures}, where tool traces or interface artifacts appear in the final output. These diagnostics indicate that retrieval invocation alone is not a sufficient measure of SF: after retrieval, the model must still determine whether the returned content is relevant, integrate it into the task solution, and preserve the required output format. Appendix~\ref{app:transition_annotation} provides the full harmful-transition taxonomy, and Appendix~\ref{app:adherence_annotation} reports the adherence-label distributions and label-conditioned RAE analysis.

\subsection{Is Negative RAE Solely an Artifact of Defective Retrieved Content?}\label{sec:skill_content_control}
Finally, we test whether negative RAE can be attributed solely to the quality of the retrieved skill text. We run skill-content controls for the two strongest coding reversal models, DeepSeek-V3.2 and Gemini-2.5-Flash-lite, replacing the normal skill returns with empty, filler, random, or corrupted returned-content conditions. Table~\ref{tab:content_controls} summarizes the pooled RAE values, while Appendix~\ref{app:content_controls} reports the corresponding raw transition counts, retrieval coverage, OAE, and aggregate retrieval lift for each control condition.

\paragraph{Returned Content Alone Does Not Determine RAE}
The content-control results indicate that negative RAE is not solely an artifact of poorly written skill content. For both DeepSeek-V3.2 and Gemini-2.5-Flash-lite, RAE remains negative across all non-empty returned-content conditions, including Normal skills, Random-skills returns, Corrupted content, and Filler-Dummy text. Thus, a structured skill artifact does not by itself guarantee a positive actual-use effect. RAE instead reflects the full retrieval-conditioned execution chain, including retrieval invocation, content injection, relevance assessment, skill integration, and final-answer formatting. These controls therefore bound, rather than fully identify, the source of negative RAE.

\section{Conclusion}\label{sec:6}
As LLMs increasingly operate within external skill environments, standard evaluation metrics—which rely on whole-benchmark averages or self-selected task subsets—create a misleading illusion of tool-use proficiency. To isolate the true causal impact of retrieved knowledge, we introduced the RAE. By strictly pairing skill-enabled and skill-disabled executions on the exact tasks where retrieval occurs, RAE neutralizes task-selection bias and directly measures the efficacy of the entire SF chain. Our evaluations across coding and mathematical domains reveal a pervasive paradox: models frequently exhibit positive aggregate retrieval lift while suffering a negative RAE. Apparent benchmark-level gains often mask actual performance degradation on the specific tasks where skills were actively deployed. By decoupling mere context exposure from actual cognitive integration, RAE provides a necessary methodological lens. It ensures that as agents become more autonomous, their tool-use mechanisms are rigorously evaluated as genuine cognitive aids rather than deceptive liabilities.

\section*{Limitations}\label{sec:limit}
Our experiments cover a focused evaluation regime: single-pass coding and mathematical tasks under a fixed retrieval interface and fixed procedural skill libraries. The results should therefore not be read as establishing the same prevalence or magnitude for long-horizon agents, executable code libraries, language-centric skill settings, natural-language annotator-based tasks, or systems with iterative memory and planning. RAE is also bounded in its statistical and causal interpretation. Because $S_{\mathrm{call}}$ is selected by the skill-enabled execution itself, RAE is not an unbiased causal effect over a pre-treatment task population, nor a model-level skill-use ability score. It is a protocol-conditional paired outcome signal: under a fixed deployment protocol, when the agent entered the retrieval-returned portion of the SF chain, did the chain help or harm the same tasks? Finer causal decomposition would require additional ablations such as oracle retrieval, metadata-only retrieval, explicit full-skill fetch, multiple skill-pool constructions, and manually adjudicated causal labels.

\section*{Ethical Considerations}
This work does not involve human subjects, private user data, or demographic profiling; all experiments are conducted on coding and mathematical benchmarks. The main ethical relevance of the study is evaluation reliability. Skill- or tool-augmented agents may be presented as improved systems when aggregate metrics hide cases where retrieval-invoked skill use harms the same tasks it was meant to help. Such overstatement can lead to misleading deployment claims and weaker auditability of agent systems. RAE is intended to make this failure mode more visible, but it should not be interpreted as a safety guarantee or as a complete causal explanation of failures. Positive or negative RAE values remain specific to the evaluated protocol, model, retrieval policy, skill library, and benchmark, and should be reported alongside complementary diagnostics and reproducibility artifacts. AI assistant tools were used to assist with coding, script debugging, result organization, and language polishing; all experimental design decisions, analyses, and final manuscript content were reviewed and controlled by the authors.

\section*{Acknowledgments}
This work was supported by the Korea Internet \& Security Agency (KISA) grant funded by the Korea government (PIPC) (No. RS-2026-25526342, Development of Technologies for Preventing Sensitive Information Inference and Risk Assessment in Foundation Model Operations). This work was also supported by the National Research Foundation of Korea (NRF) grant funded by the Korea government (MSIT) (RS-2026-25483747). This research was further supported by the Culture, Sports and Tourism R\&D Program through the Korea Creative Content Agency, funded by the Ministry of Culture, Sports and Tourism in 2026 (Project Name: Develop AI Agent Technology to Connect Knowledge through Public Cultural Facility-Based Discussion and Communication, Project Number: RS-2026-25520645).

\bibliography{custom}

\clearpage

\appendix

\newtcolorbox{promptbox}[1]{%
  enhanced jigsaw, breakable,
  colback=gray!10, colframe=gray!55,
  colbacktitle=black!80, coltitle=white,
  fonttitle=\bfseries\small\sffamily,
  title={#1}, arc=1.8mm, boxrule=0.4pt, titlerule=0pt,
  left=4pt, right=4pt, top=3pt, bottom=4pt,
  before skip=8pt, after skip=8pt,
}

\definecolor{codegreen}{rgb}{0,0.6,0}
\definecolor{codegray}{rgb}{0.5,0.5,0.5}
\definecolor{codepurple}{rgb}{0.58,0,0.82}
\definecolor{backcolour}{rgb}{0.95,0.95,0.92}

\lstdefinestyle{mystyle}{
    backgroundcolor=\color{backcolour},
    commentstyle=\color{codegreen},
    keywordstyle=\color{magenta},
    numberstyle=\tiny\color{codegray},
    stringstyle=\color{codepurple},
    basicstyle=\ttfamily\footnotesize,
    breakatwhitespace=false,
    breaklines=true,
    captionpos=b,
    keepspaces=true,
    numbers=left,
    numbersep=5pt,
    showspaces=false,
    showstringspaces=false,
    showtabs=false,
    tabsize=2
}

\lstset{style=mystyle}

\newtcolorbox{skillbox}[1]{%

  enhanced jigsaw, breakable,

  width=\linewidth,

  colback=gray!10, colframe=gray!55,

  colbacktitle=black!80, coltitle=white,

  fonttitle=\bfseries\small\sffamily,

  title={#1}, arc=1.8mm, boxrule=0.4pt, titlerule=0pt,

  left=4pt, right=4pt, top=3pt, bottom=4pt,

  before skip=8pt, after skip=8pt,

}

\section{Metric Computation and Reporting Filters}\label{app:metric_details}
All cell-level metrics are computed from pooled raw counts rather than from means of per-run rates. A cell is defined by a fixed model, benchmark, partition seed, skill pool or control condition, and evaluation protocol. When config-identical reruns exist for the same cell, we first sum the underlying counts and then compute the reported rate.

For RAE, we aggregate raw counts across config-identical runs:
\begin{equation}
    \begin{aligned}
        B_{\mathrm{cond}} &= \sum_i \mathrm{cond}_{b,i}, &
        C_{\mathrm{cond}} &= \sum_i \mathrm{cond}_{c,i}, \\
        N_{\mathrm{cond}} &= \sum_i n_{\mathrm{cond},i}.
    \end{aligned}
\end{equation}

We report
\begin{equation}
    \begin{aligned}
        \mathrm{RAE} = \Delta_{\mathrm{cond}}
        =
        100 \times
        \frac{B_{\mathrm{cond}}-C_{\mathrm{cond}}}{N_{\mathrm{cond}}}.
    \end{aligned}
\end{equation}

Here $B_{\mathrm{cond}}$ counts helpful transitions, where the skill-disabled execution fails and the skill-enabled execution succeeds, and $C_{\mathrm{cond}}$ counts harmful
transitions, where the skill-disabled execution succeeds and the skill-enabled execution fails, restricted to tasks where retrieval was invoked and at least one skill was returned in
the skill-enabled execution.

For any task subset $S$, the paired binary outcomes are
$(y_t^{+},y_t^{-})\in\{0,1\}^2$, where 1 denotes a correct final answer and 0 denotes an incorrect final answer. The four paired-outcome cells can be written as
\[
\begin{array}{c|cc}
& y_t^{+}=1 & y_t^{+}=0 \\
\hline
y_t^{-}=1 & a & c \\
y_t^{-}=0 & b & d
\end{array}
\]
where $a,b,c,d$ are task counts. The discordant cells are $b$ and $c$: $b$ counts helpful transitions, $(y_t^{+},y_t^{-})=(1,0)$,
and $c$ counts harmful transitions, $(y_t^{+},y_t^{-})=(0,1)$.

The paired outcome difference over $S$ is
\begin{equation}
    \begin{aligned}
        \frac{1}{|S|}\sum_{t\in S}(y_t^{+}-y_t^{-}).
    \end{aligned}
\end{equation}
Concordant pairs contribute zero, while helpful and harmful transitions contribute $+1$ and $-1$, respectively. Therefore,

\begin{equation}
    \begin{aligned}
        \sum_{t\in S}(y_t^{+}-y_t^{-})=b-c.
    \end{aligned}
\end{equation}

Thus, the paired same-task difference over any subset $S$ is $(b-c)/|S|$.

For OAE, we aggregate
\begin{equation}
\begin{aligned}
B_{\mathrm{all}} &= \sum_i \mathrm{all}_{b,i}, &
C_{\mathrm{all}} &= \sum_i \mathrm{all}_{c,i}, \\
N_{\mathrm{all}} &= \sum_i n_{\mathrm{pairs},i}.
\end{aligned}
\end{equation}
We report
\begin{equation}
\mathrm{OAE}
=
\Delta_{\mathrm{all}}
=
100 \times
\frac{B_{\mathrm{all}}-C_{\mathrm{all}}}{N_{\mathrm{all}}}.
\end{equation}

The decomposition into retrieval-invoked and skipped subsets follows from
$S_{\mathrm{all}}=S_{\mathrm{call}}\cup S_{\mathrm{skip}}$:
\begin{equation}
\mathrm{OAE}
=
\frac{N_{\mathrm{cond}}}{N_{\mathrm{all}}}\mathrm{RAE}
+
\frac{N_{\mathrm{skip}}}{N_{\mathrm{all}}}\Delta_{\mathrm{skip}},
\end{equation}
where
\begin{equation}
\begin{aligned}
\Delta_{\mathrm{skip}}
&=
100 \times
\frac{B_{\mathrm{skip}}-C_{\mathrm{skip}}}{N_{\mathrm{skip}}}, \\
B_{\mathrm{skip}}
&=
B_{\mathrm{all}}-B_{\mathrm{cond}}, \\
C_{\mathrm{skip}}
&=
C_{\mathrm{all}}-C_{\mathrm{cond}}.
\end{aligned}
\end{equation}

When $N_{\mathrm{skip}}=0$, the skipped-subset term is omitted.

Unlike RAE, this quantity averages the paired skill-enabled/skill-disabled difference over the entire evaluated task set, including
tasks where the agent never retrieved a skill.

For aggregate retrieval lift, let $R_i$ and $\bar{R}_i$ denote the retrieved and non-retrieved task subsets in run $i$. We aggregate
\begin{equation}
\begin{aligned}
R_{\mathrm{corr}} &= \sum_i \mathrm{correct}^{+}(R_i), &
R_N &= \sum_i |R_i|, \\
S_{\mathrm{corr}} &= \sum_i \mathrm{correct}^{+}(\bar{R}_i), &
S_N &= \sum_i |\bar{R}_i|.
\end{aligned}
\end{equation}
When both denominators are nonzero, we report
\begin{equation}
\Delta_{\mathrm{agg}}
=
100 \times
\left(
\frac{R_{\mathrm{corr}}}{R_N}
-
\frac{S_{\mathrm{corr}}}{S_N}
\right).
\end{equation}

\begin{table*}[t]
\centering
\small
\setlength{\tabcolsep}{5pt}
\begin{tabular}{lcccc}
\toprule
\textbf{Threshold} & \textbf{MBPP+ s42} & \textbf{MBPP+ s43} & \textbf{MBPP+ s44} & \textbf{HumanEval+ s42} \\
\midrule
$n_{\mathrm{cond}}\ge 5$  & $5/13$ & $6/13$ & $3/13$ & $3/13$ \\
$n_{\mathrm{cond}}\ge 10$ & $5/13$ & $6/13$ & $3/13$ & $3/13$ \\
$n_{\mathrm{cond}}\ge 15$ & $5/13$ & $6/13$ & $3/13$ & $3/12$ \\
$n_{\mathrm{cond}}\ge 20$ & $5/13$ & $6/13$ & $3/13$ & $3/11$ \\
\bottomrule
\end{tabular}
\caption{Sensitivity of aggregate-vs-RAE sign-disagreement counts to the retrieval-invoked reporting threshold. Each cell reports sign-disagreement cells divided by reportable cells. The main text uses $n_{\mathrm{cond}}\ge 10$.}
\label{tab:threshold_sensitivity}
\end{table*}

If either $R_N=0$ or $S_N=0$, aggregate retrieval lift is undefined because one of the two skill-enabled task populations is empty. This occurs, for
example, when the agent invokes retrieval on every task.

\begin{table*}[t]
\centering
\small
\setlength{\tabcolsep}{4pt}
\resizebox{\textwidth}{!}{%
\begin{tabular}{l cccc r}
\toprule
\textbf{Model}
& \textbf{MBPP+ s42}
& \textbf{MBPP+ s43}
& \textbf{MBPP+ s44}
& \textbf{HumanEval+ s42}
& \textbf{Sign disagree / cells} \\
\midrule
Qwen2.5-7B & -- & -- & -- & -- & -- \\
Qwen3-8B & $-26.4$ / $-1.5$ & $-0.5$ / $-9.2$ & $-26.6$ / $-1.6$ & \textbf{$-10.5$ / $+5.7$} & $1/4$ \\
Qwen3.5-9B & $-11.8$ / $+0.0$ & \textbf{$+26.1$ / $-3.2$} & $-36.4$ / $-4.1$ & \textbf{$-5.4$ / $+7.2$} & $2/4$ \\
Llama-3.1-8B & -- & -- & -- & -- & -- \\
\midrule
Mistral-Nemo & -- & -- & -- & -- & -- \\
Gemini-2.5-Flash-lite & \textbf{$+6.2$ / $-15.6$} & \textbf{$+16.2$ / $-15.9$} & \textbf{$+4.4$ / $-22.2$} & $-0.1$ / $-21.1$ & $3/4$ \\
Gemini-2.0-Flash & -- & -- & -- & -- & -- \\
Mistral-S-24B & $-27.2$ / $-22.5$ & \textbf{$+10.8$ / $-9.0$} & $-37.9$ / $-45.5$ & $-13.9$ / $-38.1$ & $1/4$ \\
\midrule
GPT-4o-mini & $-24.0$ / $-5.5$ & \textbf{$+11.4$ / $-3.4$} & $-23.2$ / $-8.2$ & $-2.5$ / $-1.4$ & $1/4$ \\
Claude-3.5-H & \textbf{$-5.0$ / $+2.5$} & $-1.0$ / $-2.1$ & $-6.6$ / $-3.8$ & $-2.7$ / $-7.7$ & $1/4$ \\
Claude-4.5-H & $+1.4$ / $+1.4$ & $+21.5$ / $+0.0$ & $+48.0$ / $+2.7$ & $+4.5$ / $+0.0$ & $0/4$ \\
Qwen3-32B & \textbf{$-16.9$ / $+1.8$} & $-27.9$ / $-6.0$ & $-12.4$ / $-8.5$ & $-11.2$ / $-30.1$ & $1/4$ \\
\midrule
GLM4-32B & $-25.1$ / $+0.0$ & \textbf{$+1.8$ / $-1.4$} & $-13.9$ / $-6.8$ & $-2.8$ / $-5.6$ & $1/4$ \\
Command-R & $-6.2$ / $-14.5$ & $-36.1$ / $-24.1$ & $-33.3$ / $-26.7$ & $-5.8$ / $+0.0$ & $0/4$ \\
\midrule
Llama-3.3-70B & $+69.8$ / $+4.2$ & \textbf{$+44.2$ / $-2.0$} & $+60.0$ / $+7.5$ & $+61.5$ / $+7.1$ & $1/4$ \\
Qwen3-235B & \textbf{$-0.1$ / $+7.0$} & $-21.1$ / $+0.0$ & \textbf{$-27.1$ / $+3.7$} & \textbf{$-7.0$ / $+15.5$} & $3/4$ \\
DeepSeek-V3.2 & \textbf{$+31.6$ / $-9.1$} & $-11.5$ / $-15.4$ & \textbf{$+46.2$ / $-23.1$} & $-17.1$ / $-23.0$ & $2/4$ \\
\midrule
\textbf{Sign disagree / both defined}
& $\mathbf{5/13}$ & $\mathbf{6/13}$ & $\mathbf{3/13}$ & $\mathbf{3/13}$ & \\
\textbf{Reporting cells}
& $13$ & $13$ & $13$ & $13$ & \\
\bottomrule
\end{tabular}%
}
\caption{
Full 17-model coding-domain sign-disagreement panel. Each cell reports aggregate retrieval lift / RAE in percentage points. Bold entries indicate opposite signs between the two metrics. Dashes indicate non-reportable cells, typically because the retrieval-invoked subset is empty, below the reporting threshold, or aggregate retrieval lift is undefined. Non-reportable cells are retained for transparency but excluded from sign-disagreement denominators.
}
\label{tab:full_signflip_panel}
\end{table*}

For main sign-disagreement counts, a cell is included only when $N_{\mathrm{cond}}\ge 10$ and both RAE and aggregate retrieval lift
are defined. Cells below this retrieval-invocation threshold are omitted from prevalence counts because the conditional same-task estimate is too sparse.
Cells with $N_{\mathrm{cond}}\ge 10$ but undefined aggregate retrieval lift retain their RAE value, but they are not counted in
aggregate-vs-RAE sign-disagreement denominators.

A sign disagreement is counted when the aggregate retrieval lift and RAE have opposite signs:

\begin{equation}
    \Delta_{\mathrm{agg}} \cdot \Delta_{\mathrm{cond}} < 0
\end{equation}

This criterion is metric-level. It should not be confused with task-level helpful and harmful transitions, which are the paired skill-enabled/skill-disabled outcome events aggregated by RAE.

We use $n_{\mathrm{cond}}\ge 10$ as the main reporting cutoff. Table~\ref{tab:threshold_sensitivity} shows that the aggregate-vs-RAE sign-disagreement counts remain stable under alternative cutoffs of 5, 15, and 20.

\begin{table}[t]
\centering
\small
\setlength{\tabcolsep}{1pt}
\begin{tabular}{lrrrrrr}
\toprule
\textbf{Model} & \textbf{Agg. lift} & \textbf{OAE} & \textbf{RAE} & \textbf{b} & \textbf{c} & $
\boldsymbol{n_{\mathrm{cond}}}$ \\
\midrule
Qwen2.5-7B     & --      & $-65.8$ & --      & 0  & 0  & 0 \\
Qwen3-8B       & $+0.0$  & $-10.0$ & $-20.0$ & 1  & 4  & 15 \\
Qwen3.5-9B     & $+0.7$  & $-4.2$  & $-7.1$  & 0  & 1  & 14 \\
Llama-3.1-8B   & $+5.8$  & $-10.4$ & $-9.3$  & 16 & 31 & 162 \\
Mistral-Nemo   & $-25.6$ & $-5.0$  & $+0.0$  & 0  & 0  & 2 \\
Gemini-2.5-Flash-lite  & $+13.2$ & $-18.3$ & $-11.0$ & 4  & 16 & 109 \\
Gemini-2.0-Flash & $-7.3$ & $-1.7$ & $+0.0$ & 1 & 1 & 8 \\
Mistral-S-24B  & $-64.7$ & $-1.7$  & $-20.0$ & 0  & 1  & 5 \\
GPT-4o-mini    & $+31.4$ & $+1.7$  & $+0.0$  & 0  & 0  & 1 \\
Claude-3.5-H   & $-24.7$ & $-6.2$  & $-6.5$  & 0  & 2  & 31 \\
Claude-4.5-H   & $-4.8$  & $-1.7$  & $-1.1$  & 4  & 5  & 88 \\
Qwen3-32B      & $-7.5$  & $-9.6$  & $-9.7$  & 7  & 19 & 124 \\
GLM4-32B       & $+16.6$ & $-18.3$ & $-14.4$ & 9  & 36 & 187 \\
Command-R      & $+0.8$  & $-10.4$ & $-5.5$  & 11 & 19 & 145 \\
Llama-3.3-70B  & $+14.2$ & $-42.1$ & $-39.4$ & 3  & 42 & 99 \\
Qwen3-235B     & $-3.2$  & $+2.1$  & $+0.0$  & 0  & 0  & 24 \\
DeepSeek-V3.2  & $-8.7$  & $-8.3$  & $-11.0$ & 5  & 22 & 154 \\
\bottomrule
\end{tabular}
\caption{Full Math500 cross-domain results. Values are in percentage points except for $b$, $c$, and $n_{\mathrm{cond}}$.
Aggregate retrieval lift is undefined when there are no retrieved tasks or no non-retrieved tasks for comparison. Rows with very small
$n_{\mathrm{cond}}$ are reported for completeness but are interpreted as sparse conditional estimates.}
\label{tab:math500_full_results}
\end{table}

\subsection{Representative Reversal Statistics}\label{app:key_reversal_stats}
Table~\ref{tab:key_reversal_stats} reports paired transition counts and uncertainty for representative coding-domain reversal cells. Confidence intervals use 20,000 task-level paired multinomial bootstrap resamples, and $p$-values are from exact McNemar tests over the discordant counts.

\begin{table*}[t]
\centering
\small
\setlength{\tabcolsep}{5pt}
\begin{tabular}{llrrrrll}
\toprule
\textbf{Model} & \textbf{Benchmark} & $\boldsymbol{b}$ & $\boldsymbol{c}$ & $\boldsymbol{n_{\mathrm{cond}}}$ & \textbf{RAE} & \textbf{95\% CI} & \textbf{Exact McNemar $p$} \\
\midrule
Mistral-S-24B & MBPP+ (partition 3) & 1 & 31 & 66 & $-45.5$ & $[-57.6,-33.3]$ & $1.54\!\times\!10^{-8}$ \\
DeepSeek-V3.2 & MBPP+ (partition 3) & 1 & 19 & 78 & $-23.1$ & $[-33.3,-12.8]$ & $4.01\!\times\!10^{-5}$ \\
Gemini-2.5-Flash-lite & HumanEval+ & 1 & 9 & 38 & $-21.1$ & $[-36.8,-7.9]$ & $0.0215$ \\
Qwen3-235B & HumanEval+ & 13 & 2 & 71 & $+15.5$ & $[+5.6,+25.4]$ & $0.00739$ \\
\bottomrule
\end{tabular}
\caption{Representative coding-domain reversal cells. RAE and confidence intervals are in percentage points. $b$ counts SD-wrong$\rightarrow$SE-correct transitions, $c$ counts SD-correct$\rightarrow$SE-wrong transitions, and $n_{\mathrm{cond}}$ is the retrieval-returned paired subset size.}
\label{tab:key_reversal_stats}
\end{table*}

\section{Full Coding-Domain Results}\label{app:full_coding_results}
Table~\ref{tab:full_signflip_panel} reports the full 17-model coding-domain panel underlying Table~\ref{tab:main_results}. The main table includes reportable cells only, whereas this appendix table retains all evaluated models for transparency. Dashes indicate non-reportable cells, typically because the retrieval-invoked subset is empty, below the reporting threshold, or one of the aggregate retrieval lift denominators is undefined. These cells are retained in the full panel but excluded from sign-disagreement denominators.

\section{Math500 Cross-Domain Results}\label{app:math500}

Table~\ref{tab:math500_full_results} reports the full Math500 cross-domain replication.
The table includes aggregate retrieval lift, OAE, and RAE, together with the retrieval-invoked transition counts
used to compute RAE.

\begin{table*}[t]
    \centering
    \small
    \setlength{\tabcolsep}{5pt}
    \begin{tabular}{lrrrr}
        \toprule
        \textbf{Metric} & \textbf{n} & \textbf{Raw Pearson $r$} & \textbf{Partial $r$} & \textbf{Single-control range} \\
        \midrule
        Aggregate retrieval lift & 13 & $+0.746$ & $+0.763$ & $+0.745$--$+0.755$ \\
        OAE & 13 & $+0.700$ & $+0.830$ & $+0.732$--$+0.825$ \\
        RAE & 13 & $+0.634$ & $+0.776$ & $+0.633$--$+0.747$ \\
        \bottomrule
    \end{tabular}
    \caption{Descriptive cross-benchmark correlations between MBPP+ and HumanEval+. The partial $r$ column controls for both MBPP+ and HumanEval+ skill-disabled accuracies. The
    single-control range reports the minimum and maximum partial correlations obtained when controlling for only one benchmark's skill-disabled accuracy. These values are reported as diagnostic context, not as evidence that RAE is the most cross-benchmark-stable metric.}
    \label{tab:cross_benchmark_corr}
\end{table*}

\begin{table*}[t]
\centering
\small
\begin{tabular}{lrrl}
\toprule
\textbf{Excluded model} & \textbf{MBPP+ $n_{\mathrm{cond}}$} & \textbf{HE+ $n_{\mathrm{cond}}$} & \textbf{Reason} \\
\midrule
Gemini-2.0-Flash & 0 & 0 & no retrieval-invoked conditional subset \\
Llama-3.1-8B & 0 & 0 & no retrieval-invoked conditional subset \\
Mistral-Nemo & 0 & 0 & no retrieval-invoked conditional subset \\
Qwen2.5-7B & 0 & 0 & no retrieval-invoked conditional subset \\
\bottomrule
\end{tabular}
\caption{Models excluded from the filtered cross-benchmark correlation analysis. The reporting filter requires $n_{\mathrm{cond}}\ge 10$ on both MBPP+
and HumanEval+.}
\label{tab:cross_benchmark_exclusions}
\end{table*}

\begin{table*}[t]
\centering
\small
\setlength{\tabcolsep}{4pt}
\begin{tabular}{lrrrrrr}
\toprule
\textbf{Model} & \textbf{Seed} & \textbf{Invocation} & \textbf{Retrieval} & \textbf{Adherence} & \textbf{Effective} & \textbf{RAE} \\
\midrule
DeepSeek-V3.2 & 42 & $100.0$ & $100.0$ & $18.8$ & $11.2$ & $-9.1$ \\
Gemini-2.5-Flash-lite & 42 & $82.5$  & $82.5$  & $13.6$ & $10.6$ & $-15.6$ \\
DeepSeek-V3.2 & 43 & $100.0$ & $100.0$ & $15.0$ & $11.2$ & $-15.4$ \\
Gemini-2.5-Flash-lite & 43 & $82.5$  & $82.5$  & $16.7$ & $10.6$ & $-15.9$ \\
DeepSeek-V3.2 & 44 & $97.5$  & $97.5$  & $19.2$ & $12.8$ & $-23.1$ \\
Gemini-2.5-Flash-lite & 44 & $78.8$  & $78.8$  & $17.5$ & $12.7$ & $-22.2$ \\
\bottomrule
\end{tabular}
\caption{Stage-diagnostic values for the two anchor reversal models across MBPP+ seeds. Stage values are percentages. RAE is reported in percentage points. The adherence and effective columns are diagnostic proxies from the skill-enabled logs, whereas RAE is computed from paired skill-enabled/skill-disabled outcomes.}
\label{tab:stage_diagnostic_values}
\end{table*}

\section{Cross-Benchmark Correlation Details}\label{app:cross_benchmark_corr}
Table~\ref{tab:cross_benchmark_corr} reports a descriptive cross-benchmark correlation analysis comparing MBPP+ seed 42 against HumanEval+ seed 42 under the same coding skill pool.

We include models with sufficient retrieval-invocation coverage on both benchmarks, requiring $n_{\mathrm{cond}}\ge 10$ for both MBPP+ and HumanEval+. This yields 13 models. Gemini-2.0-Flash, Llama-3.1-8B, Mistral-Nemo, and Qwen2.5-7B are excluded from this filtered correlation because at least one benchmark has fewer than 10 retrieval-invoked tasks. Table~\ref{tab:cross_benchmark_exclusions} lists these exclusions and the corresponding retrieval-invoked subset sizes.

Skill-disabled accuracy is used as the baseline ability control. For each benchmark cell, skill-disabled accuracy is reconstructed from the paired identity
\begin{equation}
    \mathrm{Acc}^{-}
    =
    \mathrm{Acc}^{+}
    -
    \mathrm{OAE}/100.
\end{equation}
Partial correlations are computed by residualizing the MBPP+ and HumanEval+ metric values against the specified skill-disabled accuracy controls and then
correlating the residuals.

In this rerun, aggregate retrieval lift has the highest raw cross-benchmark correlation, and OAE has the highest skill-disabled-accuracy-controlled partial correlation. We therefore use these correlations only to bound interpretation: RAE is a protocol-conditional actual-use signal over the agent's retrieval-invoked subset, not a benchmark-invariant ranking of model skill-use ability.

\section{Stage-Diagnostic Definitions and Values}\label{app:stage_diagnostics}
The stage diagnostics in this appendix are descriptive proxies computed from the skill-enabled execution logs and paired outcome counts. They are not used to define RAE. Their purpose is to support the main-text observation that high invocation or retrieval coverage does not imply a positive same-task actual-use outcome.

We use the following stage proxies. \textbf{Invocation} is the percentage of tasks where the agent called the skill-search interface at least once. \textbf{Retrieval} is the percentage of tasks where at least one skill was returned. \textbf{Adherence} is a lexical/structural proxy indicating whether at least one retrieved-skill identifier or code-relevant pattern appears in the submitted answer. \textbf{Effective} is the percentage of retrieval-invoked tasks where this adherence proxy is present and the final answer is correct. Table~\ref{tab:stage_diagnostic_values} reports the resulting stage-diagnostic values for the two anchor reversal models across MBPP+ seeds. The final column reports RAE, the paired same-task outcome metric over the retrieval-invoked subset.

\begin{table}[t]
\centering
\small
\setlength{\tabcolsep}{4pt}
\begin{tabular}{llrr}
\toprule
\textbf{Model} & \textbf{Primary label} & \textbf{n} & \textbf{Share} \\
\midrule
DeepSeek-V3.2 & appropriate adherence & 6 & 13.3\% \\
DeepSeek-V3.2 & format/interface failure & 7 & 15.6\% \\
DeepSeek-V3.2 & ignored/independent & 29 & 64.4\% \\
DeepSeek-V3.2 & misapplied/overapplied & 3 & 6.7\% \\
\midrule
Gemini-2.5-Flash-lite & appropriate adherence & 4 & 10.0\% \\
Gemini-2.5-Flash-lite & format/interface failure & 1 & 2.5\% \\
Gemini-2.5-Flash-lite & ignored/independent & 29 & 72.5\% \\
Gemini-2.5-Flash-lite & misapplied/overapplied & 5 & 12.5\% \\
Gemini-2.5-Flash-lite & unclear & 1 & 2.5\% \\
\bottomrule
\end{tabular}
\caption{Full harmful-transition primary-label summary before the category compression used in the main text. The unit is a skill-disabled-correct/
skill-enabled-wrong retrieval-invoked execution.}
\label{tab:harmful_taxonomy_full}
\end{table}

\begin{table}[t]
\centering
\small
\setlength{\tabcolsep}{4pt}
\begin{tabularx}{\columnwidth}{lX}
\toprule
\textbf{Failure mode} & \textbf{Definition} \\
\midrule
Appropriate adherence & The skill-enabled answer visibly uses relevant retrieved-skill content or patterns in a plausible way. \\
Ignored/independent & The skill-enabled answer retrieves a skill but shows no clear uptake of the returned content. \\
Misapplied/overapplied & The skill-enabled answer uses retrieved-skill content but applies it in a way that changes or overconstrains the task solution. \\
Format/interface failure & The skill-enabled answer fails at the output or benchmark-interface level, such as leaking tool traces, omitting the required
function, or returning malformed code. \\
Unclear & The annotator cannot assign a stable narrower behavioral label from the available evidence. \\
\bottomrule
\end{tabularx}
\caption{Primary failure-mode definitions used for harmful-transition taxonomy.}
\label{tab:harmful_taxonomy_definitions}
\end{table}

\begin{table}[t]
\centering
\footnotesize
\setlength{\tabcolsep}{2.5pt}
\begin{tabular}{@{}lccc@{}}
\toprule
Model & $b$ & $c$ & $n_{\mathrm{cond}}$ \\
\midrule
DeepSeek-V3.2 & 8 & 45 & 233 \\
Gemini-2.5-Flash-lite & 5 & 40 & 196 \\
\bottomrule
\end{tabular}
\caption{
Annotation-count reconciliation. $b$ and $c$ are helpful and harmful retrieval-invoked transitions; counts match the RAE computation.
}
\label{tab:adherence_reconciliation}
\end{table}

\begin{table*}[t]
  \centering
  \small
  \setlength{\tabcolsep}{8pt}
  \begin{tabular}{lllr}
  \toprule
  \textbf{Model} & \textbf{Transition} & \textbf{Primary label} & \textbf{Share} \\
  \midrule
  DeepSeek-V3.2 & Helpful & appropriate adherence & 50.0\% \\
  DeepSeek-V3.2 & Helpful & ignored/independent & 50.0\% \\
  DeepSeek-V3.2 & Harmful & appropriate adherence & 13.3\% \\
  DeepSeek-V3.2 & Harmful & format/interface failure & 15.6\% \\
  DeepSeek-V3.2 & Harmful & ignored/independent & 64.4\% \\
  DeepSeek-V3.2 & Harmful & misapplied/overapplied & 6.7\% \\
  \midrule
  Gemini-2.5-Flash-lite & Helpful & ignored/independent & 100.0\% \\
  Gemini-2.5-Flash-lite & Harmful & appropriate adherence & 10.0\% \\
  Gemini-2.5-Flash-lite & Harmful & format/interface failure & 2.5\% \\
  Gemini-2.5-Flash-lite & Harmful & ignored/independent & 72.5\% \\
  Gemini-2.5-Flash-lite & Harmful & misapplied/overapplied & 12.5\% \\
  Gemini-2.5-Flash-lite & Harmful & unclear & 2.5\% \\
  \bottomrule
  \end{tabular}
  \caption{GPT-5.5 adherence annotation distribution over discordant retrieval-invoked transitions. The annotator is blinded to the skill-disabled answer and
  correctness labels.}
  \label{tab:adherence_label_distribution}
\end{table*}

\begin{table*}[t]
  \centering
  \small
  \setlength{\tabcolsep}{4pt}
  \begin{tabular}{llrrrr}
  \toprule
  \textbf{Model} & \textbf{Collapsed label} & \textbf{b} & \textbf{c} & $\boldsymbol{n}$ & \textbf{RAE} \\
  \midrule
  DeepSeek-V3.2 & appropriate & 4 & 6 & 89 & $-2.2$ \\
  DeepSeek-V3.2 & no clear & 4 & 31 & 129 & $-20.9$ \\
  DeepSeek-V3.2 & problematic & 0 & 8 & 15 & $-53.3$ \\
  \midrule
  Gemini-2.5-Flash-lite & appropriate & 0 & 4 & 26 & $-15.4$ \\
  Gemini-2.5-Flash-lite & no clear & 5 & 31 & 152 & $-17.1$ \\
  Gemini-2.5-Flash-lite & problematic & 0 & 5 & 18 & $-27.8$ \\
  \bottomrule
  \end{tabular}
  \caption{Label-conditioned RAE values using collapsed GPT-5.5 adherence labels. Values are in percentage points. These diagnostic
  subsets do not redefine RAE; they describe which behavioral labels account for helpful and harmful transitions in the retrieval-
  invoked representative results.}
  \label{tab:adherence_label_conditioned_delta}
\end{table*}

\begin{table*}[t]
    \centering
    \small
    \setlength{\tabcolsep}{4pt}
    \begin{tabular}{llrrrrrrr}
        \toprule
        \textbf{Model} & \textbf{Condition} & \textbf{b/c} & $\boldsymbol{n_{\mathrm{cond}}}$ & \textbf{RAE} &
        \textbf{OAE} & \textbf{Coverage} & \textbf{Agg. lift} \\
        \midrule
        DeepSeek-V3.2 & Normal       & $8/45$  & 233 & $-15.9$ & $-15.8$ & $97.1$ & $+21.2$ \\
        DeepSeek-V3.2 & Schema-Empty & $0/0$   & 0   & --      & $-10.0$ & $0.0$  & -- \\
        DeepSeek-V3.2 & Filler-Dummy & $7/30$  & 226 & $-10.2$ & $-10.8$ & $94.2$ & $+28.0$ \\
        DeepSeek-V3.2 & Random-skills & $5/41$  & 238 & $-15.1$ & $-15.4$ & $99.2$ & $+7.1$ \\
        DeepSeek-V3.2 & Corrupted    & $8/29$  & 230 & $-9.1$  & $-9.6$  & $95.8$ & $+43.5$ \\
        \midrule
        Gemini-2.5-Flash-lite & Normal       & $5/40$ & 196 & $-17.9$ & $-17.9$ & $81.7$ & $+7.8$ \\
        Gemini-2.5-Flash-lite & Schema-Empty & $0/0$  & 0   & --      & $-15.4$ & $0.0$  & -- \\
        Gemini-2.5-Flash-lite & Filler-Dummy & $5/14$ & 181 & $-5.0$  & $-7.1$  & $75.4$ & $+8.1$ \\
        Gemini-2.5-Flash-lite & Random-skills & $2/32$ & 195 & $-15.4$ & $-17.9$ & $81.2$ & $+14.0$ \\
        Gemini-2.5-Flash-lite & Corrupted    & $5/36$ & 180 & $-17.2$ & $-17.5$ & $75.0$ & $+1.1$ \\
        \bottomrule
    \end{tabular}
    \caption{Detailed skill-content control results pooled over MBPP+ seeds 42--44. Values for RAE, OAE, coverage, and aggregate retrieval lift are in percentage points. Coverage is the percentage of tasks where retrieval returned at least one skill. Aggregate retrieval lift is undefined when retrieved or non-retrieved task populations are empty.}
    \label{tab:content_control_details}
\end{table*}

\section{Transition-Level Annotation Details}\label{app:transition_annotation}
Table~\ref{tab:harmful_taxonomy_full} reports the full harmful-transition taxonomy computed over skill-disabled-correct/skill-enabled-wrong retrieval-invoked executions from the DeepSeek-V3.2 and Gemini-2.5-Flash-lite coding-skill runs. The unit is a log-level paired execution, not a unique task. The anchor set includes MBPP+ partition seeds 42, 43, and 44 for each model, so the counts are used as qualitative diagnostic evidence rather than as an independent estimate of task-frequency prevalence.

Each harmful transition receives one primary failure-mode label. When multiple symptoms are present, the primary label is assigned to the most proximate
visible cause of the submitted answer failure. Table~\ref{tab:harmful_taxonomy_definitions} defines the labels. Table~\ref{tab:harmful_taxonomy_full} reports the full label distribution before the category compression used in the main-text taxonomy.

\section{Adherence Annotation Details}\label{app:adherence_annotation}
We annotate the pooled DeepSeek-V3.2 and Gemini-2.5-Flash-lite retrieval-invoked representative results with a GPT-5.5 annotator. The annotator receives the task prompt, retrieval query, retrieved skill names and excerpts, and the skill-enabled answer. It does not receive the skill-disabled answer or the paired correctness labels. The annotation is therefore used as diagnostic evidence about skill uptake behavior, not as part of the definition of RAE.

\begin{figure}[t]
  \centering
  \small
  \begin{subfigure}[t]{\linewidth}
      \centering
      \includegraphics[width=0.95\linewidth]{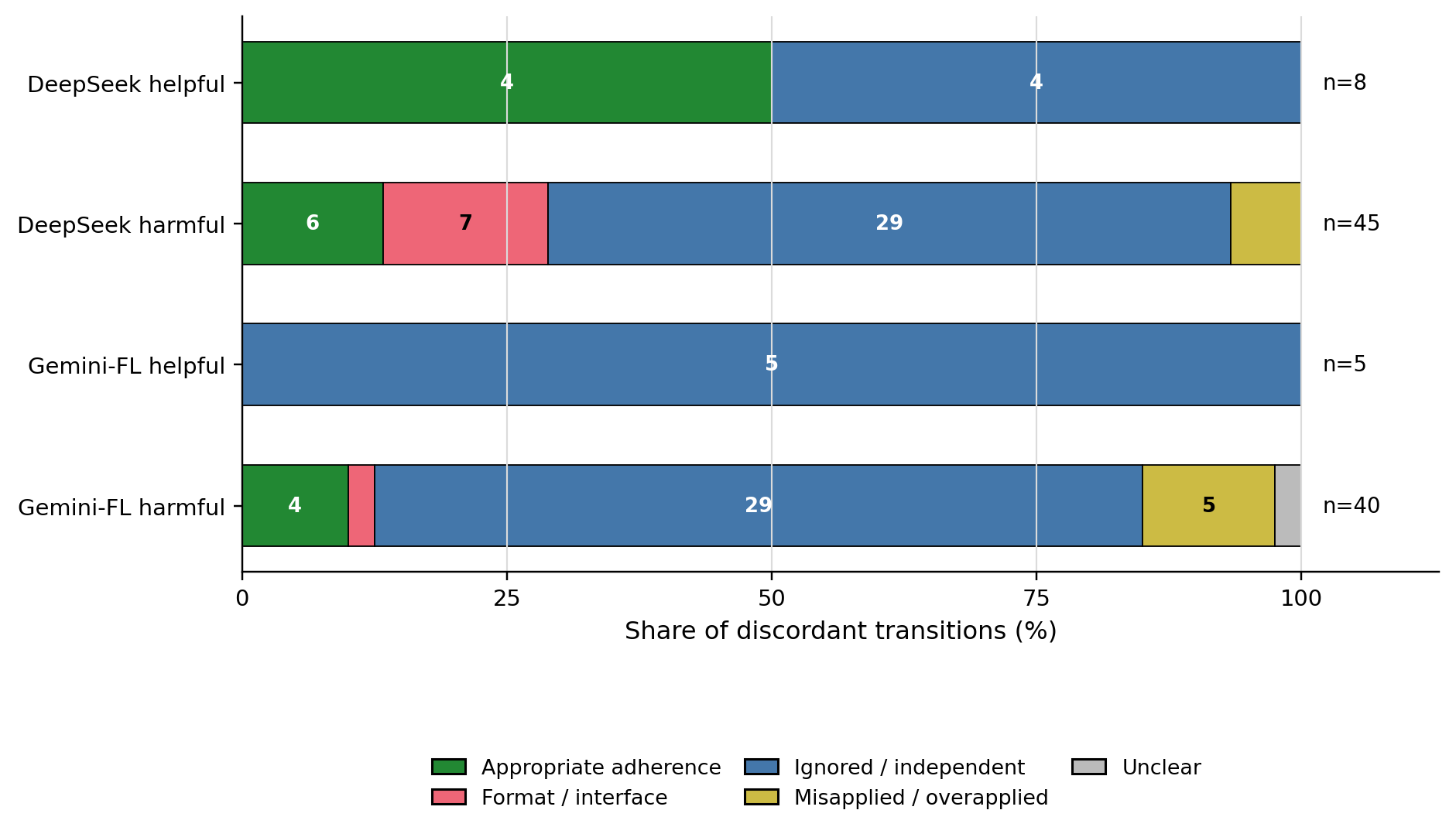}
      \caption{GPT-5.5 adherence labels over helpful and harmful transitions.}
      \label{fig:adherence_panel_a}
  \end{subfigure}

  \vspace{0.6em}

  \begin{subfigure}[t]{\linewidth}
      \centering
      \includegraphics[width=0.95\linewidth]{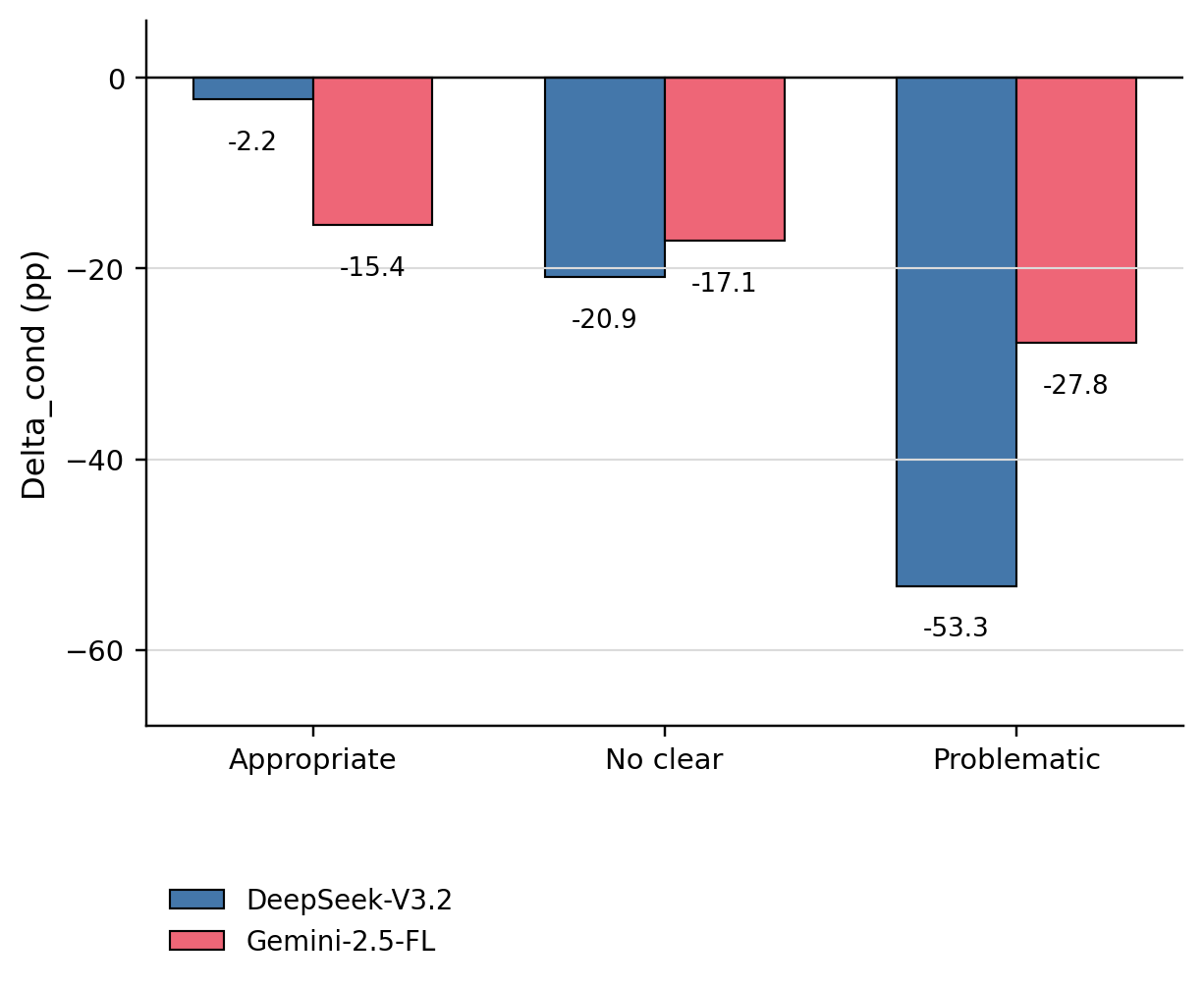}
      \caption{Label-conditioned RAE over retrieval-invoked executions.}
      \label{fig:adherence_panel_b}
  \end{subfigure}

  \caption{Adherence-conditioned diagnosis of retrieval-invoked outcome changes in the pooled DeepSeek-V3.2 and Gemini-2.5-Flash-lite runs. Panel (a) shows how helpful and harmful transitions are distributed across GPT-5.5 behavioral labels. Panel (b) reports RAE after grouping executions by collapsed adherence labels.}
  \label{fig:app_adherence_conditioned}
\end{figure}

Each skill-enabled answer receives one primary behavioral label: appropriate adherence, ignored/independent, misapplied/overapplied, format/interface failure, or unclear.
For Figure~\ref{fig:app_adherence_conditioned}, we also collapse these into three groups: appropriate adherence, no clear adherence, and problematic adherence.
The collapsed labels are used only for visualization.
Table~\ref{tab:adherence_reconciliation} reconciles the annotated logs with the transition counts used in RAE.
Table~\ref{tab:adherence_label_distribution} reports the GPT-5.5 label distribution over discordant transitions, and Table~\ref{tab:adherence_label_conditioned_delta} reports RAE after grouping retrieval-invoked executions by collapsed adherence labels.

\subsection{Human Audit of Annotation Reliability}\label{app:human_audit}

We evaluate annotation reliability on a stratified sample of 50 of the 85 harmful transitions.
Nine non-author annotators provide 150 labels while blinded to the GPT-5.5 label, the SD answer, the paired correctness labels, and one another's annotations.
Table~\ref{tab:human_audit_agreement} summarizes the main agreement results.

The following is an English translation of the full Korean instructions provided to the human annotators.

\begin{promptbox}{\texttt{Human Annotator Instructions}}
\scriptsize
\begin{verbatim}
Human audit labeling guide
(16--17 cases per person; 20--35 minutes)

Purpose. This audit tests whether the behavioral
labels assigned by an LLM (GPT-5.5) are reliable.
Relabel your cases independently; we will later
compute agreement with the LLM labels. The LLM
labels are hidden. Do not consult other annotators.

Each row contains one coding-task record:
- task_prompt: the problem given to the model
- retrieval_queries: the model's skill-search queries
- skills_retrieved / skill_text_excerpt: the skill
  documents actually returned to the model
- on_answer: the model's final answer

Read the problem, retrieved skills, and final answer,
in that order. Ask: "How did this answer handle the
retrieved skills?"

Enter exactly one label in human_primary_label:
- adhered_appropriate: visibly and appropriately
  uses retrieved-skill content for the task
- misapplied_or_overapplied: uses a retrieved skill
  incorrectly or unnecessarily
- ignored_or_independent: shows no trace of using
  the retrieved skill and solves independently
- format_or_interface_failure: fails in output or
  interface format, such as omitting the required
  function or returning prose instead of code
- unclear: cannot be assigned to the four labels
  above from the available evidence

Rules.
1. Do not judge task correctness. All cases are
   skill-enabled executions that failed; label only
   how the answer handled the retrieved skill.
2. If uncertain, choose unclear rather than forcing
   a narrower label.
3. Work alone and only on your assigned sheet.
   Notes in human_notes are optional.

Enter confidence in human_confidence_1_to_5, from
1 (guess) to 5 (certain). Low confidence is natural
for unclear cases.

Submission. Fill human_primary_label and
human_confidence_1_to_5 in your assigned
audit_annotator_A#.csv file, save it, and return it.
\end{verbatim}
\end{promptbox}

\begin{table*}[t]
\centering
\small
\begin{tabularx}{\textwidth}{Yr}
\toprule
\textbf{Audit statistic} & \textbf{Result} \\
\midrule
Five-way Krippendorff's $\alpha$ & $0.377$ \\
Five-way human-majority--GPT agreement & $52.4\%$ ($22/42$) \\
Cases without a strict five-way majority & $8/50$ \\
Binary \textit{engaged}/\textit{not engaged} human--human agreement & $84\%$ \\
Binary \textit{engaged}/\textit{not engaged} human-majority--GPT agreement & $84\%$ ($41/49$) \\
Binary Krippendorff's $\alpha$ & $0.401$ \\
Human / GPT \textit{appropriate adherence} rate & $9.3\%$ / $10.0\%$ \\
\bottomrule
\end{tabularx}
\caption{Human-audit agreement for the GPT-5.5 annotations. The five labels are \textit{appropriate adherence}, \textit{ignored/independent}, \textit{misapplied/overapplied}, \textit{format/interface failure}, and \textit{unclear}. For the binary analysis, \textit{appropriate adherence} and \textit{misapplied/overapplied} are grouped as \textit{engaged}, while \textit{ignored/independent}, \textit{format/interface failure}, and \textit{unclear} are grouped as \textit{not engaged}.}
\label{tab:human_audit_agreement}
\end{table*}

Agreement is modest under the original five-way taxonomy, indicating that the fine-grained category boundaries are interpretive.
The binary \textit{engaged}/\textit{not engaged} distinction is more stable, with $84\%$ human--human and human-majority--GPT agreement.
We therefore use the labels as descriptive post-hoc diagnostics and limit our conclusion to the low prevalence of \textit{appropriate adherence} among the audited harmful transitions.

\section{Skill-Content Control Details}\label{app:content_controls}
Table~\ref{tab:content_control_details} reports the pooled raw counts underlying Table~\ref{tab:content_controls}. The three Normal cells use distinct 80-task partitions, whereas the control replicates reuse the same 80-task partition; pooled arm values are therefore descriptive. On shared tasks, the Gemini Normal--Filler RAE differences are $-7.7$, $-14.0$, and $-3.8$ pp across the three control replicates, with only the second interval excluding zero. DeepSeek Normal--Filler differences are $+4.1$, $+4.1$, and $0.0$ pp. These heterogeneous comparisons do not identify a single content mechanism. Schema-Empty has $n_{\mathrm{cond}}=0$ because the interface returns no usable skill, so RAE is undefined while OAE remains defined over the whole benchmark.

\section{Retrieval-Configuration Sensitivity}\label{app:retrieval_sensitivity}
We rerun three representative models on the same 80 MBPP+ tasks with retrieval depths $k\in\{1,3,5\}$ and with a mechanically merged single-skill library. All runs use one generic agent role, temperature $0.7$, maximum output length 8,192, reasoning disabled, and at most three agent turns. Confidence intervals use 20,000 paired multinomial bootstrap resamples; $p$-values are within-configuration exact McNemar tests.

\begin{table*}[t]
\centering
\footnotesize
\setlength{\tabcolsep}{3.5pt}
\begin{tabular}{llrrrrrll}
\toprule
\textbf{Model} & \textbf{Configuration} & $\boldsymbol{b}$ & $\boldsymbol{c}$ & $\boldsymbol{n_{\mathrm{cond}}}$ & \textbf{Coverage} & \textbf{RAE} & \textbf{95\% CI} & \textbf{Exact $p$} \\
\midrule
Gemini-2.5-FL & $k=1$ & 0 & 9 & 65 & $81.2$ & $-13.8$ & $[-23.1,-6.2]$ & $0.00391$ \\
 & $k=3$ (paper) & 2 & 12 & 64 & $80.0$ & $-15.6$ & $[-26.6,-4.7]$ & $0.0129$ \\
 & $k=5$ & 1 & 9 & 70 & $87.5$ & $-11.4$ & $[-20.0,-2.9]$ & $0.0215$ \\
 & merged-1 & 0 & 2 & 23 & $28.7$ & $-8.7$ & $[-21.7,+0.0]$ & $0.500$ \\
\midrule
DeepSeek-V3.2 & $k=1$ & 0 & 19 & 77 & $96.2$ & $-24.7$ & $[-35.1,-15.6]$ & $3.81\!\times\!10^{-6}$ \\
 & $k=3$ (paper) & 3 & 10 & 77 & $96.2$ & $-9.1$ & $[-18.2,+0.0]$ & $0.0923$ \\
 & $k=5$ & 4 & 11 & 77 & $96.2$ & $-9.1$ & $[-19.5,+0.0]$ & $0.118$ \\
 & merged-1 & 4 & 6 & 71 & $88.8$ & $-2.8$ & $[-11.3,+5.6]$ & $0.754$ \\
\midrule
Claude-4.5-H & $k=1$ & 1 & 2 & 72 & $90.0$ & $-1.4$ & $[-6.9,+2.8]$ & $1.000$ \\
 & $k=3$ (paper) & 3 & 2 & 70 & $87.5$ & $+1.4$ & $[-4.3,+7.1]$ & $1.000$ \\
 & $k=5$ & 4 & 4 & 70 & $87.5$ & $+0.0$ & $[-8.6,+7.1]$ & $1.000$ \\
 & merged-1 & 0 & 3 & 19 & $23.8$ & $-15.8$ & $[-31.6,+0.0]$ & $0.250$ \\
\bottomrule
\end{tabular}
\caption{Retrieval-depth and merged-library sensitivity. Coverage is the percentage of tasks for which at least one skill was returned; RAE and confidence intervals are in percentage points. The merged-1 intervention changes both skill granularity and retrieval coverage, so it is not interpreted as a pure granularity effect.}
\label{tab:retrieval_sensitivity}
\end{table*}

Across $k=1,3,5$, Gemini remains negative with significant within-configuration tests, DeepSeek remains negative with significance only at $k=1$, and Claude remains near zero. The repeated direction for Gemini and DeepSeek rules out the default $k=3$ choice as the sole explanation, while the merged-1 results remain jointly affected by lower coverage.

\section{Skill Pool Details}\label{app:skill_pools}
Tables~\ref{tab:coding_skill_pool} and~\ref{tab:math_skill_pool} summarize the fixed procedural skill pools used in the coding and mathematical
evaluations. Skills are stored as Markdown files following the AgentSkill-style structure described in \S\ref{sec:4.2}. The retrieval index uses the
skill descriptions and Markdown bodies, and the top retrieved skill texts are appended to the agent context when retrieval is invoked.

\begin{table*}[t]
\centering
\small
\setlength{\tabcolsep}{4pt}
\begin{tabularx}{\columnwidth}{lX}
\toprule
\textbf{Coding skill} & \textbf{Summary} \\
\midrule
binary-search-boundary & Find a threshold in a sorted or monotonic search space. \\
counting-hash & Count occurrences with a hash map for frequency, duplicate, or anagram-style tasks. \\
dfs-iterative & Traverse graphs iteratively for reachability, connected components, or cycle-safe search. \\
edge-case-guard & Add defensive checks for empty, \texttt{None}, singleton, and boundary inputs. \\
knapsack-01-dp & Use one-dimensional bottom-up dynamic programming for 0/1 knapsack-style constraints. \\
prefix-sum & Build prefix sums for constant-time contiguous range queries. \\
rotate-left & Rotate a sequence left by a specified offset with wrap-around behavior. \\
sliding-window-longest & Find the longest contiguous window satisfying a monotonic predicate. \\
two-pointer-pair-sum & Find pair or closest-sum patterns in sorted or sortable sequences. \\
\bottomrule
\end{tabularx}
\caption{Coding skill pool used for MBPP+ and HumanEval+. The pool contains 9 procedural skills.}
\label{tab:coding_skill_pool}
\end{table*}

The Math skill pool is summarized in Table~\ref{tab:math_skill_pool}.

\begin{table*}[t]
\centering
\small
\setlength{\tabcolsep}{4pt}
\begin{tabularx}{\columnwidth}{lX}
\toprule
\textbf{Math skill} & \textbf{Summary} \\
\midrule
algebraic-substitution & Substitute known values into an expression and simplify. \\
case-analysis & Split a problem into mutually exclusive cases such as sign, parity, range, or absolute-value branches. \\
check-by-substitution & Verify a derived answer by substituting it back into the original equations or constraints. \\
coordinate-distance & Compute distance, midpoint, slope, or angle from coordinate representations. \\
factor-then-solve & Factor polynomial equations and solve by setting factors to zero. \\
modular-arithmetic & Work with remainders, divisibility, cyclic patterns, and large-power computations. \\
sequence-formulae & Apply standard closed forms for arithmetic, geometric, and telescoping sequences. \\
systematic-enumeration & Count or list candidates by traversing a bounded finite search space. \\
\bottomrule
\end{tabularx}
\caption{Math skill pool used for the Math500 cross-domain replication. The pool contains 8 procedural skills.}
\label{tab:math_skill_pool}
\end{table*}

\section{Model Panel}\label{app:models}
Table~\ref{tab:model_panel} lists the model panel used in our experiments. The coding panel consists of the 17 models (Claude 3.5 Haiku\footnote{\url{https://www-cdn.anthropic.com/c7822cdc35ad788ec87e14b3a9d45010f1f86c38.pdf}}, Claude Haiku 4.5\footnote{\url{https://www.anthropic.com/news/claude-haiku-4-5}}, Command R\footnote{\url{https://huggingface.co/CohereLabs/c4ai-command-r-v01}}, DeepSeek V3.2, Gemini 2.0 Flash\footnote{\url{https://ai.google.dev/gemini-api/docs/models/gemini-2.0-flash}}, Gemini 2.5 Flash-Lite, Llama 3.1 8B Instruct, Llama 3.3 70B Instruct, Mistral NeMo\footnote{\url{https://huggingface.co/mistralai/Mistral-Nemo-Instruct-2407}}, Mistral Small 3.2 24B Instruct\footnote{\url{https://huggingface.co/mistralai/Mistral-Small-3.1-24B-Instruct-2503}}, GPT-4o mini\footnote{\url{https://openai.com/ko-KR/index/gpt-4o-mini-advancing-cost-efficient-intelligence}}, Qwen2.5 7B Instruct, Qwen3 235B A22B 2507, Qwen3 32B, Qwen3 8B, Qwen3.5 9B, GLM4-32B\footnote{\url{https://huggingface.co/zai-org/GLM-4-32B-0414}})~\cite{liu2025deepseek, comanici2025gemini, grattafiori2024llama, yang2025qwen3, qwen3.5, qwen2.5} evaluated on MBPP+ and HumanEval+. The Math500 cross-domain replication uses the same 17-model panel, with reportability varying according to retrieval-invoked coverage. Model identifiers correspond to the API-facing model names used in the experiment logs.

\begin{table*}[t]
    \centering
    \small
    \setlength{\tabcolsep}{4pt}
    \begin{tabular}{llll}
        \toprule
        \textbf{Paper alias} & \textbf{Model identifier} & \textbf{Provider / family} & \textbf{Access type} \\
        \midrule
            Claude-3.5-H & \texttt{anthropic/claude-3.5-haiku} & Anthropic Claude & Closed/API \\
            Claude-4.5-H & \texttt{anthropic/claude-haiku-4.5} & Anthropic Claude & Closed/API  \\
            Command-R & \texttt{cohere/command-r-08-2024} & Cohere Command-R & Closed/API   \\
            DeepSeek-V3.2 & \texttt{deepseek/deepseek-v3.2} & DeepSeek & Open-weight family \\
            Gemini-2.0-Flash & \texttt{google/gemini-2.0-flash-001} & Google Gemini & Closed/API \\
            Gemini-2.5-Flash-lite & \texttt{google/Gemini-2.5-Flash-lite} & Google Gemini & Closed/API \\
            Llama-3.1-8B & \texttt{meta-llama/llama-3.1-8b-instruct} & Meta Llama & Open-weight family \\
            Llama-3.3-70B & \texttt{meta-llama/llama-3.3-70b-instruct} & Meta Llama & Open-weight family \\
            Mistral-Nemo & \texttt{mistralai/mistral-nemo} & Mistral AI & Open-weight family \\
            Mistral-S-24B & \texttt{mistralai/mistral-small-3.2-24b-instruct} & Mistral AI & Open-weight family \\
            GPT-4o-mini & \texttt{openai/gpt-4o-mini} & OpenAI GPT & Closed/API \\
            Qwen2.5-7B & \texttt{qwen/qwen-2.5-7b-instruct} & Qwen & Open-weight family \\
            Qwen3-235B & \texttt{qwen/qwen3-235b-a22b-2507} & Qwen & Open-weight family \\
            Qwen3-32B & \texttt{qwen/qwen3-32b} & Qwen & Open-weight family \\
            Qwen3-8B & \texttt{qwen/qwen3-8b} & Qwen & Open-weight family \\
            Qwen3.5-9B & \texttt{qwen/qwen3.5-9b} & Qwen & Open-weight family \\
            GLM4-32B & \texttt{z-ai/glm-4-32b} & Z-AI GLM & Open-weight family  \\
        \bottomrule
    \end{tabular}
    \caption{Model panel used in the experiments. Access type is used only to describe the diversity of the evaluated panel; all models were accessed through
    the same OpenAI-compatible experiment interface.}
    \label{tab:model_panel}
\end{table*}

\

\section{Agent Prompts and Tool Schema}\label{app:skill_and_prompts}
This appendix reports the prompts and tool schema used in the paired SE and SD executions and in the adherence annotation. The SE and SD prompts are kept identical except for the availability of the \texttt{search\_skills} interface. The annotator prompt is used only for post hoc diagnostic labeling and is not used in the computation of RAE.

\paragraph{Execution settings.} Generation uses temperature $0.7$, a maximum output length of 8,192 tokens, provider-side reasoning disabled, and at most three sequential \texttt{search\_skills} calls. Each valid call returns the top-3 BM25 matches with their full skill text. Invalid arguments return no skill and generation continues; when several calls are emitted together, only the first is executed. No manual context truncation is applied. Transient requests are retried up to three times, and unrecoverable provider or context-length failures produce an empty prediction scored as incorrect.

\paragraph{Answer extraction and correctness.} For coding tasks, the evaluator extracts the first Python-fenced block, falls back to the first generic fenced block, and otherwise evaluates the raw response. The extracted program is executed against the EvalPlus extended test suite and is correct only when every test passes. Math answers are evaluated with \texttt{math-verify}, followed by boxed-answer parsing and normalized exact matching.

\clearpage

\begin{promptbox}{\texttt{SE Agent Prompt}}
\footnotesize
\begin{verbatim}
You are a helpful assistant.

You have a `search_skills` tool that searches a
skill library.

Use it if you think it could help solve the task.

Implement the requested function
with the exact signature.
Reply inside a
```python block — only the function,
no test code.
\end{verbatim}
\end{promptbox}

\begin{promptbox}{\texttt{SD Agent Prompt}}
\footnotesize
\begin{verbatim}
You are a helpful assistant.

Implement the requested function with the
exact signature.

Reply inside a ```python block
— only the function, no test code.
\end{verbatim}
\end{promptbox}

\begin{promptbox}{\texttt{GPT-5.5 Annotator Prompt}}
\footnotesize
\begin{verbatim}
You annotate whether an LLM answer
substantively follows retrieved skill content.

Do not infer from task correctness;
no correctness labels are provided.

Return one compact JSON object only.
\end{verbatim}
\end{promptbox}

\begin{lstlisting}[language=Python, caption={Search skill tool schema}, label={lst:search_skills_tool}]
SEARCH_SKILLS_TOOL = {
    "type": "function",
    "function": {
        "name": "search_skills",
        "description": (
            "Search the skill library for reusable problem-solving artifacts. "
            "Returns utility functions (callable helpers you can use directly), "
            "procedure templates (algorithm skeletons you can specialize), and "
            "guard checklists (defensive checks to add at function entry)."
        ),
        "parameters": {
            "type": "object",
            "properties": {
                "query": {
                    "type": "string",
                    "description": "Natural-language query describing the subproblem or pattern you need.",
                },
                "types": {
                    "type": "array",
                    "items": {"type": "string", "enum": ["utility", "procedure", "guard"]},
                    "description": (
                        "Optional filter. Omit for all types. Use "
                        "['utility', 'procedure'] to skip defensive-only skills."
                    ),
                },
            },
            "required": ["query"],
        },
    },
}
\end{lstlisting}

\end{document}